%% file: main_journal_saga_arxiv.tex
\documentclass[10pt,journal,compsoc]{IEEEtran}
\usepackage{amsmath,amsfonts}
\usepackage{array}
\usepackage[caption=false,font=normalsize,labelfont=sf,textfont=sf]{subfig}
\usepackage{textcomp}
\usepackage{stfloats}
\usepackage{url}
\usepackage{verbatim}
\usepackage{graphicx}
\usepackage{cite}
\usepackage{tabularx}
\usepackage{multirow}
\usepackage{algorithmicx}
\usepackage{algorithm}
\usepackage{array}
\usepackage{algpseudocode}
\usepackage{hyperref}

\input{preamble}

\input{pkgs}

\newcommand{\x}{\mathit{\mathbf{x}}}
\newcommand{\verts}{\mathbf{v}}
\newcommand{\face}{\mathbf{f}}
\newcommand{\bary}{\boldsymbol{a}}

\newcommand{\quater}{\mathbf{q}}
\newcommand{\gaus}{\mathcal{G}}

\newcommand{\gcov}{\mathbf{\Sigma}}

\newcommand{\pose}{\boldsymbol{\theta}}
\DeclareMathOperator*{\argmin}{arg min}

\title{SAGA: Surface-Aligned Gaussian Avatar}

\author{Ronghan Chen, Yang Cong,~\IEEEmembership{Senior Member,~IEEE}, Jiayue Liu
	\thanks{
	This work is supported in part by the National Key R\&D Program of China under Grant 2023YFB4704800, and the National Natural Science Foundation of China under Grant (62225310, 62127807). (Corresponding author: Prof. Yang Cong.)
		
	Ronghan Chen is with the State Key Laboratory of Robotics, Shenyang Institute of Automation, Chinese Academy of Sciences, Shenyang 110016, China, and also with the University of Chinese Academy of Sciences, Beijing 100049, China (e-mail: louischen965@gmail.com).
	
	Yang Cong and Jiayue Liu is with the School of Automation Science and Engineering, South China University of Technology, Guangzhou 510640, China (e-mail: congyang81@gmail.com, aujiayueliu@mail.scut.edu.cn).
}
}

\begin{document}

\IEEEtitleabstractindextext{

\begin{abstract}
This paper presents a Surface-Aligned Gaussian representation for creating animatable human avatars from monocular videos, aiming at improving the novel view and pose synthesis performance while ensuring fast training and real-time rendering. 
Recently, 3D Gaussian Splatting (3DGS) has emerged as a more efficient and expressive alternative to neural radiance fields (NeRF), and has been used for creating dynamic human avatars. 
However, when applied to the severely ill-posed task of monocular dynamic reconstruction, the Gaussians tend to overfit the constantly changing regions such as clothes wrinkles or shadows since these regions cannot provide consistent supervision, resulting in \emph{noisy geometry} and \emph{abrupt deformation} that typically fail to generalize under novel views and poses. 
To address these limitations, we present SAGA, \emph{i.e.}, Surface-Aligned Gaussian Avatar, which aligns the Gaussians with a mesh to enforce \emph{well-defined geometry} and \emph{consistent deformation}, thereby improving generalization under novel views and poses. 
Unlike existing strict alignment methods that suffer from limited expressive power and low realism, SAGA employs a two-stage alignment strategy where the Gaussians are first \emph{adhered on} while then \emph{detached from} the mesh, thus facilitating both good geometry and high expressivity. 
In the first \emph{Adhered Stage}, we improve the flexibility of Adhered-on-Mesh Gaussians by allowing them to flow on the mesh, in contrast to existing methods that rigidly bind Gaussians to fixed location.
In the second \emph{Detached Stage}, we introduce a Gaussian-Mesh Alignment regularization, which allows us to unleash the expressivity by detaching the Gaussians but maintain the geometric alignment by minimizing their location and orientation offsets from the bound triangles.
Finally, since the Gaussians may drift outside the bound triangles during optimization, an efficient Walking-on-Mesh strategy is proposed to dynamically update the bound triangles, ensuring accurate regularization even as the geometry evolves. 
Experiments on challenging datasets demonstrate that SAGA outperforms both NeRF and Gaussian-based methods on novel view and pose synthesis tasks, with fast training time of \textbf{12} minutes, and real-time rendering efficiency at \textbf{60+} FPS. Additionally, we showcase that SAGA enables direct high-quality mesh extraction from Gaussians, marking the first attempt at deformable Gaussians learned from monocular human videos.

\end{abstract}

}
\maketitle
\IEEEdisplaynontitleabstractindextext
\IEEEpeerreviewmaketitle

\section{Introduction}
\begin{figure*}
	\includegraphics[width=0.99\linewidth]{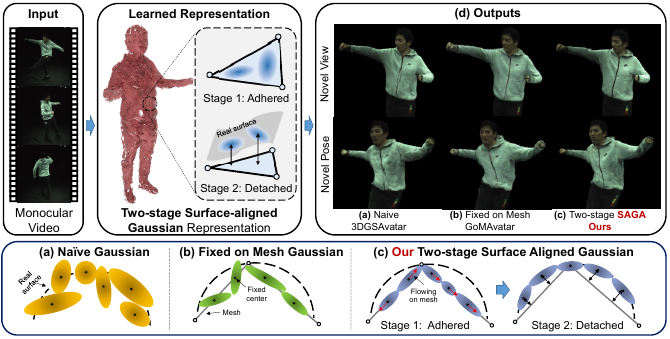}
	\caption{
		\textbf{UPPER:} Illustration of SAGA, \emph{i.e.} Surface-aligned Gaussian Avatar for monocular drivable avatar reconstruction and animation. \textbf{LOWER:} Since monocular dynamic reconstruction is severely ill-posed, state-of-the-art methods either \textbf{(a)} overfit the scene with naive Gaussians or \textbf{(b)} overconstrain the Gaussians by fixing them on the mesh. In contrast, \textbf{(c)} SAGA applies a first-adhered-then-detached manner to effectively regularize Gaussians without sacrificing the expressivity, \textbf{(d)} leading to more photorealistic rendering results. 
	}
	\label{fig:intro}
\end{figure*}
Free-view rendering of animatable humans is a challenging task with broad applications in Augmented Reality (AR), telepresence, movies and video production. Traditional methods~\cite{dou2016fusion4d, orts2016holoportation, guo2019relightables, collet2015high} reconstruct high-resolution textured mesh and estimate material properties~\cite{guo2019relightables} to achieve photorealistic rendering, which often require meticulously arranged lab setups and costly equipments, such as dense RGB and IR camera arrays~\cite{guo2019relightables, collet2015high} or 3D scanners~\cite{dou2016fusion4d, orts2016holoportation}, making these methods impractical for general consumers.

The advent of neural rendering~\cite{tewari2020state}, particularly Neural Radiance Fields (NeRF)~\cite{nerf} has revolutionized novel view synthesis, enabling photorealistic human rendering~\cite{peng2021neural, weng2022humannerf} and animation~\cite{peng2021neural, peng2024animatable, liu2021neural, jiang2022neuman} from sparse-view images. NeRF-based methods ease the traditional modeling setup and reach new levels of realism by learning a neural representation that can be optimized to minimize the render error. But they model such representation with large multi-layer perceptron (MLP), which requires extended training time ($>$ 10 hours), and cannot be rendered in real-time. While efficient neural implicit representations~\cite{instant-ngp,TensoRF,dvgo, Plenoxels,hedman2021snerg, PlenOctrees} have emerged, the efficiency of neural human reconstruction and rendering has yet not been satisfying~\cite{instant-nvr,instantavatar}, due to the high memory and computational complexity of volumetric representation. Moreover, they still struggle to fit highly dynamic motions from monocular video, leading to blurry synthesized results.

Recently, 3D Gaussian Splatting~\cite{gaussplat} (3DGS), as a point cloud-like representation, has significantly surpassed NeRF-based methods in both quality and rendering speed. Leveraging 3DGS, some recent methods~\cite{qian20233dgs, gauhuman, Zielonka2023Drivable3D, hugs, gart, zheng2024gps, li2024animatable} have demonstrated its potential for dynamic human reconstruction and rendering. However, 3D Gaussians with ultimate expressive power tend to overfit the region with inconsistent view-dependent appearance rather than form a good geometry~\cite{sugar}, leading to artifacts in distinct views~\cite{Lu_2024_CVPR, gaussianpro}. 
This problem, initially identified in static scenes, is exacerbated in monocularly captured dynamic scenes, where much severer inconsistency emerges from transient regions such as clothes wrinkles or shadows. 
Changing constantly with rapid human motion, these regions cannot provide consistent supervision for the Gaussians,
resulting in \textbf{\emph{noisy geometry}} and \textbf{\emph{abrupt deformation}} that typically fail to generalize  to novel views and poses.

On the other side of the spectrum, meshes have long been explored as a representation for human rendering and animation, offering consistent multi-view performance and well-defined geometry~\cite{habermann2020deepcap, saito2021scanimate, habermann2021real, Huang_2020_CVPR}. They are also easier to manipulate and generalize to new poses compared to Gaussians. 
To this end, some methods anchor the Gaussians on the mesh to regularize them with well-defined mesh geometry~\cite{sugar,gomavatar}. However, due to great geometry and topology discrepancy between the SMPL~\cite{smpl} mesh and the real human surface, such rigid binding significantly limits the expressive power of 3DGS, making it extremely difficult to capture the highly non-rigid deformations caused by complex human motion, resulting in severe loss of realism.

To address these challenges, we propose a new Two-Stage Surface-Aligned Gaussian representation, which aligns the Gaussians with a coarse human mesh~\cite{smpl} to enforce the Gaussians to form \textbf{\emph{well-defined geometry}} and learn \textbf{\emph{consistent deformation}}, thereby improving novel view and pose generalization ability. 
Meanwhile, we aim at maximally maintaining the expressivity of the Gaussians to ensure the rendering realism by designing a two-stage alignment strategy. We name the method SAGA, i.e., Surface-Aligned Gaussian Avatar.

Specifically, in the two stages, the Gaussians are first \emph{\textbf{adhered}} on the mesh to guide them to form a well-defined geometry, and then \emph{\textbf{detached}} from the mesh to unleash the expressivity of 3DGS for fine structures. 
In the first stage, unlike previous methods that rigidly adhere the Gaussians on the mesh~\cite{sugar, gomavatar}, we allow them to flow freely to improve the flexibility. Specifically, we learn Gaussian centers by simultaneously optimizing barycentric coordinates and the corresponding triangle vertices, which not only ensures them to be strictly on the mesh but also allows local adjustment to better fit the scene. We also align their orientation by flattening the Gaussians and setting their normals to match the triangle normals. This design ensures that the mesh regulates the Gaussians to form good geometry while the Gaussians in turn drive the SMPL mesh to quickly align with the real surface.

In the second stage, the Gaussians are detached from the mesh to fit finer structures. To maintain the geometry quality of the detached Gaussians, we propose a \emph{Gaussian-Mesh Alignment Regularization} to regularize them by minimizing the center and orientation offset from the mesh. This regularization can also serve as a deformation regularizer, since it constrains the deformed Gaussians to be bound to the same triangles across all the training frames.
Finally, during optimization, the Gaussians may drift out of the bound triangles, resulting in incorrect mesh-based regularization. Thus, we propose an efficient \emph{Walking-on-Mesh strategy} to accurately update the corresponding triangles with minimal computational overhead.

Experiments on challenging datasets show that our method produces more photorealistic results in novel view synthesis, and generalizes better in novel pose synthesis. Additionally, while existing Gaussian-based methods achieve satisfactory rendering results, they cannot extract high-quality meshes from Gaussians.
In contrast, SAGA significantly improves the geometry quality, marking the first successful attempt at direct high-quality mesh extraction from deformable Gaussians reconstructed from monocular human videos.

In summary, our main technical contributions include:
\begin{itemize}
	\item{
		We propose to leverage mesh as a geometric regularizer for Gaussians based monocular avatar reconstruction with a new two-stage surface-aligned representation, where the Gaussians are first adhered on the mesh to enforce well-defined geometry thereby preventing overfitting, and then detached to fully exploit the expressivity to fit finer details.
	}
	\item{
		An Adhered-on-Mesh representation that, contrary to existing methods that rigidly bind Gaussians to fixed locations, allows them to flow freely on the mesh with higher flexibility.
	}
	\item{
		A Gaussian-Mesh Alignment regularization that enforces the detached Gaussians to form well-defined geometry and learn more consistent deformation.
		}
	\item{
	Since the Gaussians can drift outside the bound triangle, we propose an efficient Walking-on-Mesh strategy to update the triangles, ensuring correct mesh-based regularization.
	}
	\item{Our method achieves SOTA novel view and pose synthesis results, and, to our knowledge, for the first time enables direct mesh extraction from deformable human Gaussians reconstructed from monocular videos.}
\end{itemize}

\section{Related Work}
We divide previous methods into NeRF-based human avatars, methods that leverage efficient NeRFs~\cite{instant-ngp} to accelerate training and rendering, more recent Gaussian-based human avatar and finally methods that integrating Gaussians with meshes.

\paragraph{Neural Radiance Fields based Human Avatar.}
Given the unprecedented success of Neural Radiance Fields~\cite{nerf}, many methods have applied NeRF to reconstruct and render humans from videos~\cite{peng2021neural,peng2021animatable,weng2022humannerf,liu2021neural,noguchi2021neural,su2021nerf,jiang2022neuman,xu2022surface}. Since the original NeRF cannot model dynamic human motions, these methods leverage deformation priors such as skeletons~\cite{noguchi2021neural, su2021nerf, peng2021animatable, weng2022humannerf} or SMPL~\cite{smpl,jiang2022neuman} model~\cite{peng2021neural, liu2021neural} to warp the neural fields. A line of works~\cite{peng2021animatable,weng2022humannerf,liu2021neural,jiang2022neuman,xu2022surface} build a canonical neural radiance field, and warp the points in each frame to the canonical space via inverse Linear Blend Skinning (LBS). 
AnimatableNeRF~\cite{peng2021animatable} learns a neural blend weight field. \cite{liu2021neural,jiang2022neuman} further introduce a non-rigid module to compensate detailed deformation. 
Other methods develop learnable~\cite{peng2021neural} or hand-crafted embeddings~\cite{su2021nerf} to encode the sampled points. NeuralBody~\cite{peng2021neural} anchors latent codes to the vertices of a SMPL model, and diffuses to the whole observation space via sparse convolution. A-NeRF~\cite{su2021nerf} handcrafts a skeleton-relative encoding, thus avoiding ill-posed inverse transformation.

Despite the impressive rendering quality, these methods generally take a long training time ($>$10h) to reconstruct only a \emph{single} person, and cannot render in real-time. Though some generalizable methods achieve fast finetuning on new persons, they are limited to well-calibrated multi-view setting~\cite{kwon2021neural, mps-nerf, Doublefield}.

\paragraph{Efficient Neural Human Avatar.}
To enable efficient training and rendering of dynamic human video, recent methods~\cite{instant-nvr,instantavatar} have applied InstantNGP~\cite{instant-ngp} as the human representation. InstantAvatar~\cite{instantavatar} further improves the efficiency by introducing an occupancy field to prune points from the empty space. Instant-NVR~\cite{instant-nvr} designs specific hash embedders for each human part to adjust representational power based on part complexity thus accelerating the convergence. Though reducing training time to minutes, 
they still cannot achieve real-time rendering at $\ge$ 24 FPS. 
Another line of work builds efficient representation based on shape primitives, such as meshes~\cite{deliffas}, voxels~\cite{remelli2022drivable} or patches~\cite{zheng2023avatarrex} to effectively reduce the sampled points for acceleration. However, they generally require to reconstruct a more accurate template or bake a texture map from multi-view inputs, which cannot be applied to monocular videos and require days of training.

\paragraph{Gaussian based Human Avatar.}
Since 3D Gaussian Splatting~\cite{gaussplat} achieves significant breakthrough in novel view synthesis on static scenes in terms of rendering quality and efficiency, recent trend has shifted to transferring 3DGS from static scene to dynamic scene reconstruction~\cite{luiten2023dynamic,duan20244d,das2024neural,yang2024deformable}, especially dynamic humans~\cite{Hifi4g,qian20233dgs,hu2024gaussianavatar,gauhuman,Zielonka2023Drivable3D,hugs,gart,splatarmor,zheng2024gps,li2024animatable,jiang2024uv}.

Similar to NeRF-based methods, these methods typically model Gaussians in the canonical space, and apply LBS or DQB model to warp the Gaussians to different poses. 
For methods taking multi-view videos as inputs~\cite{li2024animatable, zheng2024gps,ash,jiang2024uv}, an image-to-image translator~\cite{isola2017image} is typically used to generate 2D texture map to model high-fidelity motion-dependent textures. Moreover, Animatable Gaussian~\cite{li2024animatable} leverages a more accurate template and directly translates it to Gaussian parameters. GPS-Gaussian~\cite{zheng2024gps} proposes a generalizable NVS method, which first predicts depth map with stereo-based methods, and then regresses Gaussian parameters. Generally, the calibrated multi-view setup of these methods relatively limits their usage. Moreover, the image translator requires long training time of more than 10 hours.

For monocular based methods, 3DGS-Avatar~\cite{qian20233dgs} applies pose-dependent color and non-rigid MLP to predict color and deformation for each Gaussian in the canonical space. An as-isometric-as-possible regularization~\cite{DynPtFields} is further introduced to regularize the deformation. A similar framework is also applied in~\cite{hugs,gart,splatarmor,gauhuman}. These methods typically require minutes of training, and render at $>50$ FPS. However, due to the ill-posedness of dynamic monocular reconstruction and ultimate expressivity of Gaussians, they still suffer from overfitting, leading to undesirable artifacts under novel views and poses. 

\paragraph{Integrating Gaussians with Mesh.}
Meshes are naturally multi-view consistent with well-defined geometry and easier to manipulate, offering better generalization ability. Some methods propose to integrate Gaussians with meshes to combine their advantages~\cite{sugar, splattingavatar, gomavatar,haha}. 
SuGaR~\cite{sugar} extracts a mesh from Gaussians and re-anchors them onto the mesh for further refinement. However, this approach is not suitable for dynamic scenes, where it is more difficult to reconstruct an accurate enough mesh to anchor on. 
In the context of dynamic humans, most methods employ the parametric human model, SMPL~\cite{smpl}. For example, GoMAvatar~\cite{gomavatar} anchors Gaussians at the centers of mesh triangles. 
However, such constraints can be overly rigid and prevent the Gaussians from fitting the scene accurately, leading to loss of rendering realism. Moreover, they ignore the alignment of orientation.
SplattingAvatar~\cite{splattingavatar} defines Gaussian center as optimizable uv coordinates and distance above the triangle and does not optimize the SMPL mesh. 
It primarily focuses on leveraging the mesh to manipulate the Gaussians without providing adequate geometric regularization.
Moreover, while lifted optimization techniques~\cite{shen2020phong, taylor2014user, taylor2016efficient} are introduced to update the bound triangles, they are inefficient and prone to inaccurate triangles, thus, as we will show (Fig.~\ref{fig:peoplesnapshot}, Tab.~\ref{tab:ablation_zju_walking_on_triangles}) being 200$\times$ slower with artifacts.
In contrast, we develop a flexible Surface-aligned Gaussian representation that effectively regularizes them to form a well-defined geometry without sacrificing the fitting ability, and a Walking-on-Mesh strategy that tracks precise triangles efficiently. 

\section{Preliminary}
\noindent
\textbf{3D Gaussians.}\label{sec:3.1}
As an explicit representation, 3DGS models a static scene as a set of 3D Gaussians~\cite{gaussplat} $\{\gaus_i\}_{i=1}^N$, where each Gaussian $\gaus_i$ is defined by mean $\mathit{\x}_i\in \mathbb{R}^3$ and covariance matrix $\gcov_i\in \mathbb{R}^{3\times 3}$, and additionally assigned with opacity $\sigma_i\in \mathbb{R}^{1}$ and color $\mathbf{c}_i\in \mathbb{R}^{3}$ values:
\begin{equation}
	\gaus_i=(\x_i, \gcov_i, \sigma_i, \mathbf{c}_i).
\end{equation}
Intuitively, a 3D Gaussian is analogous to an ellipsoid. So ${\x}$ describes the center location, and $\gcov$ describes the scaling and orientation via the decomposition~\cite{gaussplat}:
\begin{equation}
	\gcov=\mathbf{RSS}^T\mathbf{R}^T,
\end{equation}
where $\mathbf{R}\in\mathbb{R}^{3\times 3}$ is the rotation matrix determining the orientation, and $\mathbf{S}\in \mathbb{R}^{3\times 3}$ is the diagonal scaling matrix defined as:
\begin{equation}
	\mathbf{S}=\operatorname{diag}(\mathbf{s}),
\end{equation}
where its diagonal entries are parameterized by the scaling vector $\mathbf{s}=[s_{0},s_{1},s_{2}]^{T}$ determining the scales along each axis.

\noindent
\textbf{Rendering 3D Gaussians.}
In contrast to sampling along rays in NeRF-based methods, 3D Gaussians~\cite{zwicker2002ewa} are rendered more efficiently by projecting them onto 2D image: $\gaus_i^{2D}=(\x^{\operatorname{2D}}_i, \gcov^{\operatorname{2D}}_i, \sigma_i, \mathbf{c}_i)$, where $\x^{\operatorname{2D}}_i$ is the projection of the 3D Gaussian mean $\x_i$ in the image, and the covariance $\gcov^{\operatorname{2D}}_i$ is given by:
\begin{equation}
	\mathbf{\Sigma}^{\operatorname{2D}}_i=\boldsymbol{JV}\mathbf{\gcov}_i \boldsymbol{V}^{T}\boldsymbol{J}^{T},
\end{equation}
where $\boldsymbol{V}$ is the viewpoint matrix, and $\boldsymbol{J}$ is the Jacobian of the projection matrix. Finally, the color of each pixel $p$ is obtained by $\alpha$-blending the 2D Gaussians that overlap it:
\begin{equation}
	C(p) = \sum_{i}^N \alpha_i \prod_{j<i}(1-\alpha_j)\mathbf{c}_i,
\end{equation}
where $\alpha_i$ is given by the opacity $\sigma_i$ multiplied by the probability of the pixel in the projected 2D Gaussian distribution.

\paragraph{Animatable 3D Gaussians.}
To reconstruct animatable 3D Gaussians, most methods decompose the dynamic human into canonical human Gaussians, and a deformation function to warp the Gaussians into different human poses. The deformation function is typically composed of articulated deformation that blends the rigid transformations of each human bone, and non-rigid deformation that models fine-grained deformation, such as clothes wrinkles or human expressions. Details of these modules are introduced in Sec.~\ref{sec:def_color}. We further propose a mesh-based regularization to regularize the non-rigid deformation which we introduce in Sec.~\ref{sec:reg}.

\section{Method}
\begin{figure*}[htbp]
	\begin{center}
		\includegraphics[width=1.0\linewidth]{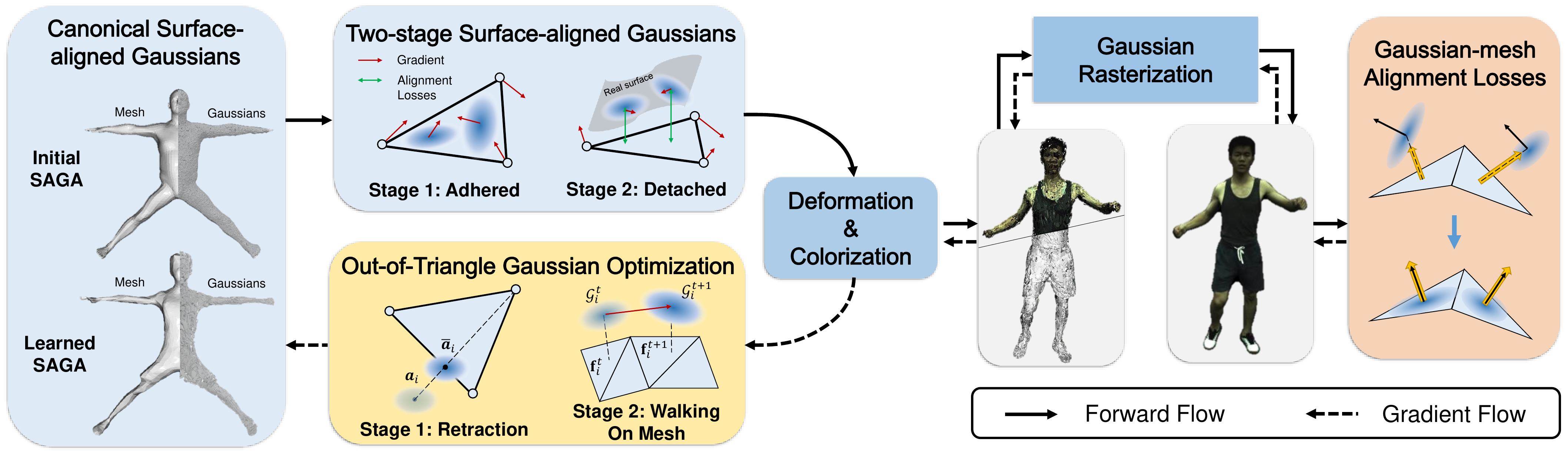}
		\caption{\textbf{The framework of Surface-aligned Gaussian Avatar (SAGA).} We model the human with a \emph{two-stage Surface-Aligned Gaussian} representation in the canonical space, where the Gaussians are first strictly adhered on the SMPL mesh (Stage 1, Sec.~\ref{sec:stage_one}), and then detached from the mesh to fit finer details (Stage 2, Sec.~\ref{sec:stage_two}). The canonical Gaussians are sent into the \emph{Deformation \& Colorization Module} to transform them to the observation space, predict the non-rigid deformation, and compensate the color changes caused by motion (Sec.~\ref{sec:def_color}).
			Finally, the Gaussians are rasterized to render the image. For backpropagation, we compute the \emph{Gaussian-Mesh Alignment Losses} to regularize the deformed Gaussians to align with the mesh in the Detached Stage (Sec.~\ref{sec:reg}). To prevent the incorrect regularization when a Gaussian moves outside the triangle, we use the proposed \emph{retraction} and \emph{Walking-on-Mesh} strategies to retract the Gaussian back within the triangle or update the new bound triangle in the first and second stages, respectively (Sec.~\ref{sec:walk}).}
		\label{fig:main}
	\end{center}
\end{figure*}

Given a monocular video of a human performer with estimated human poses and masks, our goal is to learn a Gaussian representation to rerender the video from free views, and animate the human to perform novel actions. Moreover, by exploiting the efficiency of 3DGS, we expect our method to inherit its fast training time and real-time rendering efficiency.

For monocular human reconstruction, naive Gaussians often suffer from overfitting and struggle to synthesize plausible results under novel views and poses. To address these challenges, we present SAGA, Surface-Aligned Gaussian Avatar, which leverages a template human mesh as a proxy regularizer to constrain the Gaussians to form well-defined surface that enhances generalization in novel view and pose rendering. Meanwhile, SAGA maximally maintains the expressivity of 3DGS with a flexible two-stage representation. 
As shown in Fig.~\ref{fig:main}, in stage 1, \emph{i.e.}, the \textbf{\emph{Adhered Stage}}, we adhere the Gaussians on the mesh to guide them to form a well-defined geometry (Sec.~\ref{sec:stage_one}). Then in stage 2, \emph{i.e.}, the \textbf{\emph{Detached Stage}}, we detach the Gaussians from the mesh to unleash the expressivity for fine details (Sec.~\ref{sec:stage_two}).
To prevent detached Gaussians' geometry from being corrupted and constrain the non-rigid deformation, we introduce the \emph{Gaussian-Mesh Alignment Regularization} in the detached stage (Sec.~\ref{sec:reg}).
Additionally, since the Gaussians may move outside their bound triangles during optimization, we develop a \emph{retraction strategy} for the Adhered Stage and a \emph{Walking-on-Mesh strategy} for the Detached Stage to accurately update the corresponding triangles (Sec.~\ref{sec:walk}).
Finally, we describe how the canonical Gaussians are colorized and transformed to the observation space in Sec.~\ref{sec:def_color}.

\begin{figure}[t]
	\begin{center}
		\includegraphics[width=1.0\linewidth]{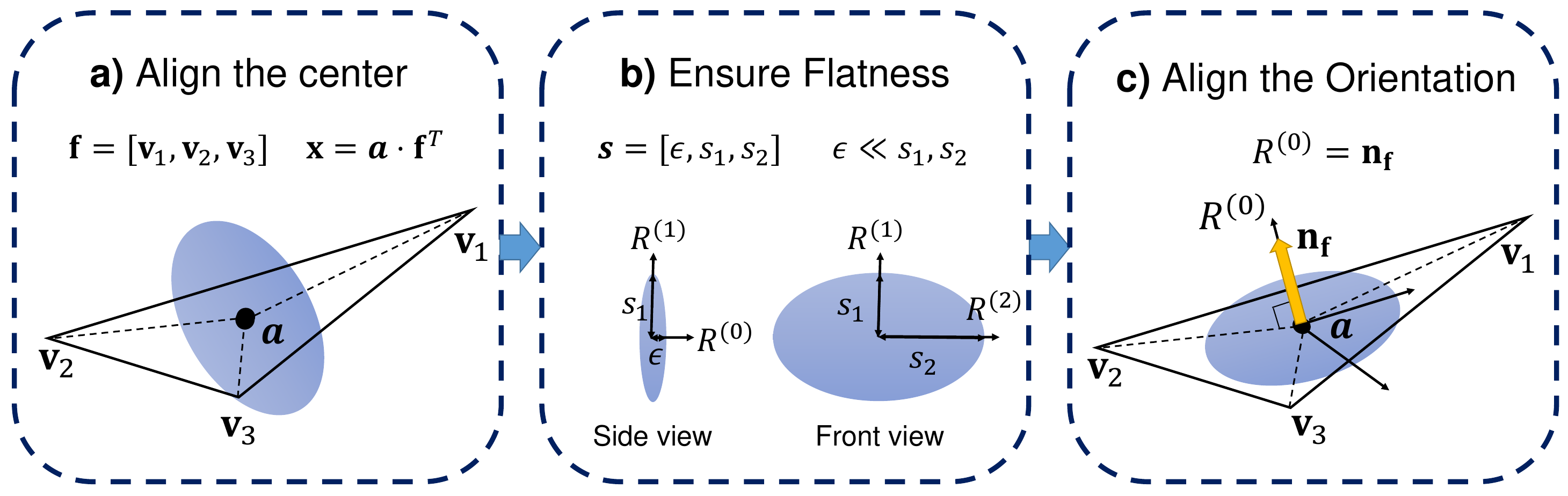}
		\caption{\textbf{Illustration of the Adhered-on-Mesh Gaussian.} We first \textbf{a)} align the Gaussian center on the triangle by defining it based on barycentric coordinates. Then in \textbf{b)}, we make the Gaussian flat, and \textbf{c)} fix the direction of the smallest scale $\mathbf{R}^{(0)}$ as the triangle normal $\mathbf{n}_{\face}$ to align the Gaussian orientation with the surface. Different from former fixed-on-mesh representation~\cite{sugar,gomavatar}, we simultaneously optimize the barycentric coordinates $\bary$ and the mesh vertices $\verts$, which allows Gaussians to flow on the mesh for higher flexibility while driving the mesh to fit the scene.}
		\label{fig:saga}
	\end{center}
\end{figure}
\subsection{Stage 1: Adhering Gaussians on the surface}\label{sec:stage_one}

As pointed out in~\cite{gaussplat}, one limitation of original 3DGS is its tendency to produce artifacts in regions that have view-dependent appearance due to inconsistent supervision. Originally found on static scenes, such limitation only becomes even severer on monocularly captured dynamic scene, where there are much more dynamic regions that change constantly following human motions.

We believe that well-defined geometry is a prior that helps to prevent such overfitting. 
Thus, we introduce the SMPL mesh as a proxy regularizer to enforce the Gaussians to form a well-defined surface. The key idea is to align the Gaussian position and orientation with the mesh. Former methods achieve this by binding one or several orderly arranged Gaussians rigidly at fixed location in each triangle~\cite{gomavatar,gao2024mesh,sugar}. Unfortunately, this ties Gaussians' density with mesh topology, and limits the expressivity.
To solve this problem, we allow the Gaussians to flow freely on the mesh during optimization.

Specifically, we represent Gaussian centers as barycentric coordinates in the bound mesh triangles, and jointly optimize these coordinates and triangle vertices to fit the scene. We use the SMPL model~\cite{smpl} as a coarse human mesh in the canonical space, which is denoted as $\mathcal{M}=(V, F)$ composed of the vertex set $V=\{\verts_i\}$ and the face set $F=\{\face_i\}$, where $\verts_i\in \mathbb{R}^3$ is the 3D vertex coordinates, and $\face_i=[\verts_{i,1},\verts_{i,2},\verts_{i,3}]\in \mathbb{R}^{3\times 3}$ is a tuple of vertices that comprise the triangle face. 

Here we take one Gaussian as an example, and omit the subscript of Gaussian index for clarity.
As illustrated in Fig.~\ref{fig:saga} \textbf{a)}, for a Gaussian $\gaus$ bound to the triangle $\face$, we define its barycentric coordinates as $\bary=[a_1, a_2, a_3]\in \mathbb{R}^{1\times 3}$. Then, the Gaussian center $\x$ is computed by interpolating the face vertices $\face$ with the barycentric coordinates $\bary$:
\begin{equation}
\x=\bary\cdot\face^{T}=a_1\verts_{1}+a_2\verts_{2}+a_3\verts_{3}.
\end{equation}

With the centers on the mesh, existing dynamic Gaussian representations~\cite{gomavatar,gao2024mesh} often ignore the alignment of orientation. As a result, the Gaussians may stick out of the surface, causing artifacts. Thus, as illustrated in Fig.~\ref{fig:saga} \textbf{b)}, we further flatten the Gaussians and, in Fig.~\ref{fig:saga} \textbf{c)}, align the orientation with the mesh. 
Specifically, to ensure flatness, we set the first Gaussian scale $s_0=\epsilon$ as a small constant value $\epsilon$ with $\epsilon\ll s_1, s_2$:
\begin{equation}
	\mathbf{s}=[\epsilon,s_1,s_2]
\end{equation}
In this way, each flat Gaussian can be regarded as a surfel, and its normal $\mathbf{n}_{\gaus}$ is the axis with the smallest scale,
represented as $R^{(0)}$. Here the index $0$ means the \emph{first} column of the rotation matrix $R$, which corresponds to the index of the axis with the smallest scale.
Then we align the Gaussians' orientation with the mesh by setting the Gaussian normal as the normal of its bound triangle $\mathbf{n}_{\face}$:
\begin{equation}
	R^{(0)}=\mathbf{n}_{\face}=(\verts_{2}-\verts_{1})\times(\verts_{3}-\verts_{1}).
\end{equation}
To this end, the original 3 degree-of-freedom (DoF) rotation is reduced to only one DoF in-plane rotation denoted as an angle $\beta$. 

In summary, the SAGA in the adhered stage can be fully parameterized by the following learnable parameters:
\begin{equation}
	\gaus=(\bary,\face,s_1,s_2, \beta).
\end{equation}

\subsection{Stage 2: Detaching for refinement}\label{sec:stage_two}
While the Adhered Stage ensures the Gaussians are well-aligned with the surface, the constraint could be overly restrictive and hinder the Gaussians to fit the scene accurately, due to the great discrepancy between the coarse SMPL mesh and real surface. 
To solve this problem, we introduce the second Detached Stage, where the strictly adhered-on-mesh constraint is relaxed, allowing Gaussians to detach from the mesh to fit finer details.

\subsubsection{Detaching and rebinding}
The \emph{detached} Gaussian representation is defined as:
\begin{equation}\label{eq:detach}
	\gaus=(\x,\mathbf{S},\mathbf{R},\bary,\face),
\end{equation}
where we remove the strict center and orientation alignment constraints by resetting the Gaussian parameters as the original form (first three terms in Eq.~\ref{eq:detach}).

We then \emph{rebind} the Gaussians to the triangles to maintain a loose connection for further mesh-based regularizations. We achieve this by keeping the last two terms in Eq.~\ref{eq:detach} from the adhered Gaussians, \emph{i.e.}, the barycentric coordinates $\bary$ and the bound triangle $\face$. 
Different from the first stage, we no longer use these terms to compute the Gaussian center but directly optimize the center $\x$ through gradient descent. Moreover, we do not optimize the barycentric coordinates $\bary$, but directly obtain it analytically by projecting the Gaussian center $\x$ on the corresponding triangle $\face$:
\begin{equation}\label{eq:proj_bary}
	\bary=\operatorname{Proj}(\x, \face),
\end{equation}
We provide the detailed derivation of this equation in the supplementary material.

After rebinding, we develop several regularizations between each Gaussian and the bound triangle to constrain the detached Gaussians, which are introduced as follows. 

\subsubsection{Gaussian-Mesh alignment regularization}\label{sec:reg}
\begin{figure}[htbp]
	\includegraphics[width=1.0\linewidth]{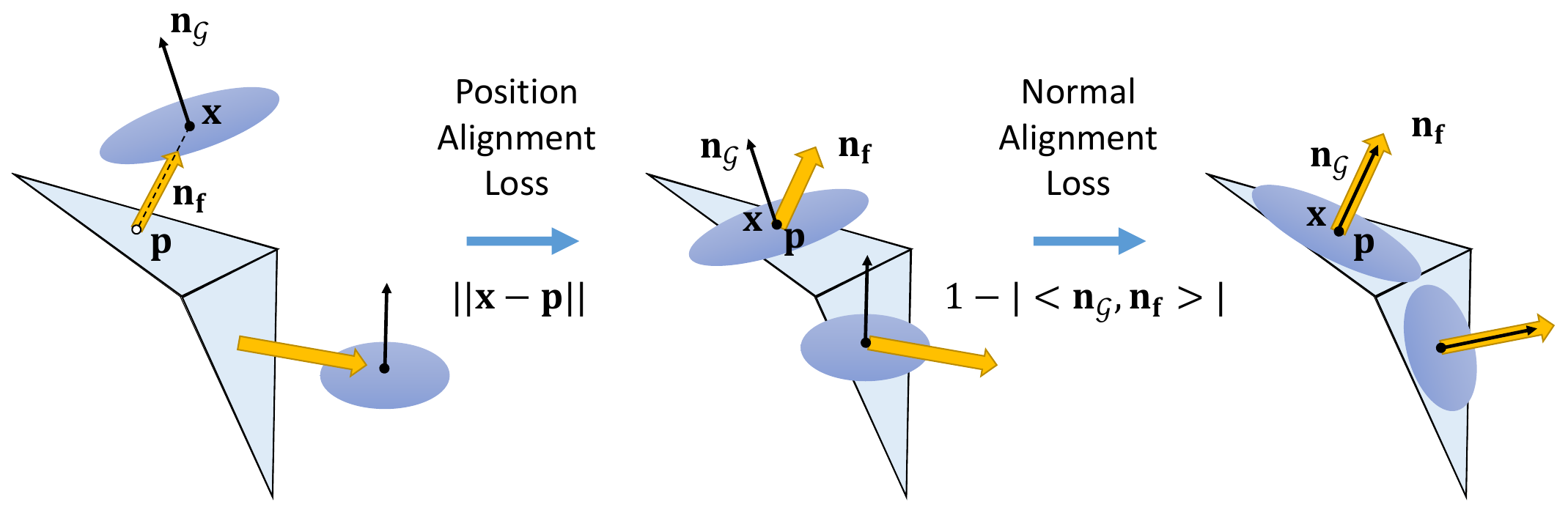}
	\caption{\textbf{Illustration of the Gaussian-mesh alignment losses}, which consists of a position and a normal alignment loss.}
	\label{fig:loss}
\end{figure}

We align the detached Gaussians' geometry with the mesh through position and orientation alignment losses, which also enforce smoother and more consistent non-rigid deformation by keeping the deformed Gaussians being bound to the same triangle across all training frames. This is illustrated in Fig.~\ref{fig:loss}.

\paragraph{Position Alignment Loss.}
For a Gaussian $\gaus$, the position-alignment loss is defined as the minimum distance between the Gaussian center $\x$ and the corresponding triangle $\face$:
\begin{equation}\label{eq:pa}
	L_{\rm pa}=\sum_{i=0}^{N} \min_{\mathbf{p}\,\operatorname{in}\,\face}\|\x-\mathbf{p}\|^2,
\end{equation}
where $\mathbf{p}$ is a point within the triangle $\face$.
Note that comparing to a naive point-to-plane distance, this loss additionally confines the Gaussians within the triangle from the tangential direction.

\paragraph{Orientation Alignment Loss.}
We align the orientation of the Gaussians with the mesh by penalizing the difference between their normals. Specifically, we use the cosine distance:
\begin{gather}\label{eq:na}
	L_{\rm na}=1-|<\mathbf{n}_{\gaus}, \mathbf{n}_{\face}>|, \\
	\mathbf{n}_{\gaus}=\mathbf{R}^{(j)}	\quad	j=\argmin_j\{s_j\}_{j=1}^3,
\end{gather}
where $|<\cdot, \cdot>|$ denotes the absolute value of the cosine similarity, and $\mathbf{n}_{\gaus}$ denotes the Gaussian normal, which is the axis with the smallest scale in the rotation matrix $\mathbf{R}$.

\noindent
\textbf{Mesh-based Regularization.}
With the above alignment losses, we can use the mesh as a proxy to transfer desired property, e.g., smoothness, from the mesh to Gaussians with sophisticated mesh-based regularizations. Here, we introduce a Laplacian-based and a normal-based mesh smoothness regularization:
\begin{equation}\label{eq:smooth}
	L_{s}=\lambda_{\rm lap} L_{\rm lap}+\lambda_{\rm normal}L_{\rm normal},
\end{equation}
where the Laplacian loss $L_{\rm lap}$ minimizes differences of adjacent vertices weighted by cotangent weights, and the normal loss $L_{\rm normal}$ minimizes the cosine distance of adjacent face normals.

\subsection{Optimizing the out-of-triangle Gaussians}\label{sec:walk}
During optimization, the Gaussians may drift outside their bound triangles. This causes the Gaussians to be aligned with wrong triangles, thus resulting in incorrect regularization that corrupts the geometry. 

To address this, we propose two strategies: \textbf{1) Retraction} that moves the out-of-triangle Gaussian back within the same triangle, which is simple but somewhat limits the flexibility, and \textbf{2) Walking-on-Mesh} that updates the bound triangle as the Gaussian moves, which is more flexible but could be expensive. 

We apply the \textbf{retraction} in the Adhered Stage for simplicity, and design an efficient \textbf{Walking-on-Mesh} strategy in the Detached Stage to maintain the expressivity for fitting finer details. Details of these two strategies are introduced as follows.

\subsubsection{Retraction}
\begin{figure}[htbp]
	\includegraphics[width=1.0\linewidth]{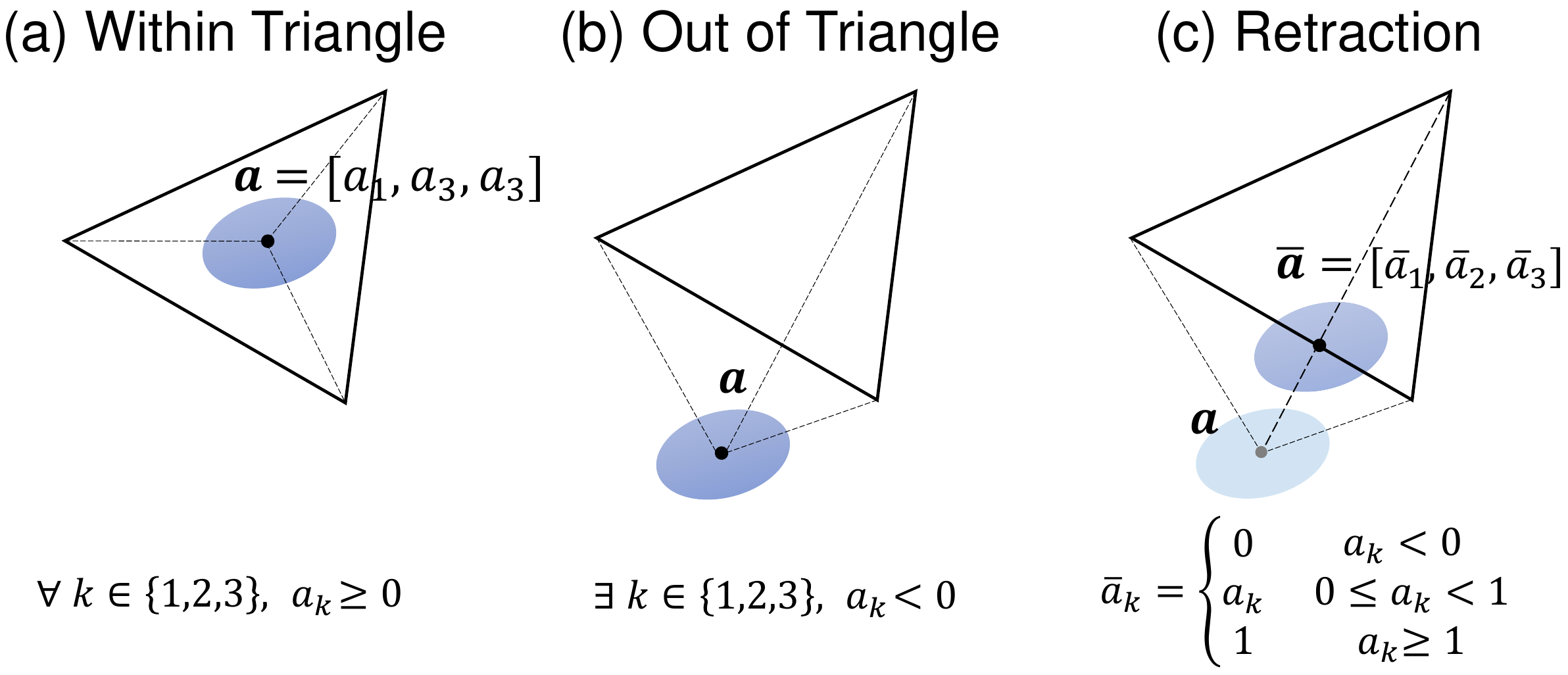}
	\caption{\textbf{Illustration of the retraction strategy} for out-of-triangle Gaussians optimization in the Adhered Stage.}
	\label{fig:overshoot}
\end{figure}

As shown in Fig.~\ref{fig:overshoot}(a)(b), we first determine if a Gaussian $\gaus$ is out of the triangle based on the barycentric coordinates $\bary=(a_1,a_2,a_3)$:
\begin{equation}\label{eq:oot}
\operatorname{OutOfTriangle}(\bary)=
\begin{cases}
	0		&	\forall k\in\{1,2,3\}, \frac{a_k}{a_1+a_2+a_3}\geq0\\
	1		&	\exists k\in\{1,2,3\}, \frac{a_k}{a_1+a_2+a_3}<0,
\end{cases}
\end{equation}
which means that a Gaussian is inside the triangle if all the normalized barycentric coordinates $a_k$ are $\geq0$, and is outside the triangle if one of the normalized barycentric coordinates $a_k$ is $<0$.
Then, we retract the out-of-triangle Gaussian back on the closest edge in Fig.~\ref{fig:overshoot}(c), via:
\begin{equation}
	\bar{a}_k=
	\begin{cases}
		0		&	a_k\leq0\\
		a_k	&	0< a_k\leq1\\
		1		&	a_k > 1
	\end{cases},\quad	
	k \in \{1,2,3\}
\end{equation}
where $\bar{a}_k$ denotes the retracted barycentric coordinates.

\subsubsection{Walking-on-Mesh strategy}
\begin{figure}[htbp]
	\includegraphics[width=1.0\linewidth]{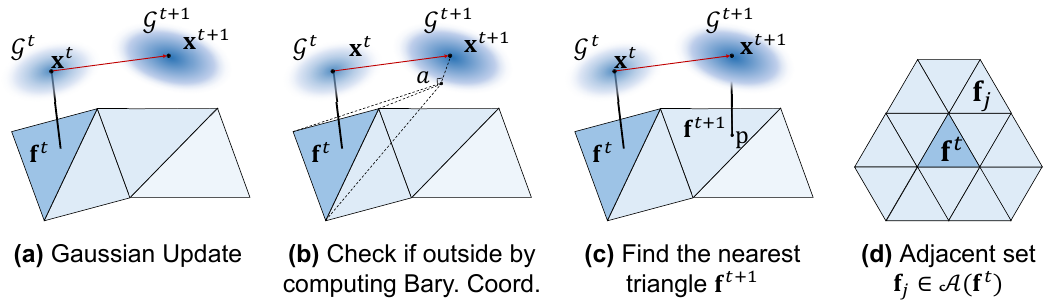}
	\caption{\textbf{Illustration of the Walking-on-Mesh strategy} applied in the Detached Stage. \textbf{(a)} An optimization step updates Gaussian $\gaus^t$ bound with triangle $\face^t$ to $\gaus^{t+1}$. \textbf{(b)} We first check whether the updated Gaussian $\gaus^{t+1}$ is out of the current triangle $\face^t$  based on projected bary. coord. $\bary$ (Eq.~\ref{eq:oot}). \textbf{(c)} If it is, we update its bound triangle as the closest one in the adjacent set $\mathcal{A}(\face^t)$ shown in (d).}
	\label{fig:walk}
\end{figure}
In the Detached Stage, we update the bound triangle as a Gaussian drifts outside the current one by proposing a Walking-on-Mesh algorithm.

As illustrated in Fig.~\ref{fig:walk}, given a Gaussian $\gaus^t$ centered at $\x^t$ and bound to triangle $\face^t$, an optimization step updates the Gaussian position to $\x^{t+1}$. The Walking-on-Mesh algorithm finds the new triangle $\face^{t+1}$ that the Gaussian should be bound to. We define the new triangle $\face^{t+1}$ as the nearest triangle to the new center $\x^{t+1}$ based on Euclidean distance. However, naively finding the closest triangle for tens of thousands of Gaussians on the whole human mesh with $>130,000$ triangles is expensive and significantly decreases the training speed.

Thus, we \textbf{first} reduce the number of walking Gaussians by only considering the out-of-triangle Gaussians. As shown in Fig~\ref{fig:walk}(b), we project the new center $\x^{t+1}$ back to the current face $f^{t}$ and obtain the barycentric coordinates $\bary$ (Eq.~\ref{eq:proj_bary}). Then we determine whether it is outside via Eq~\ref{eq:oot}.

\textbf{Secondly}, we reduce the search scope to a local triangle set $\mathcal{A}(\face^t)$ that is adjacent to the current triangle $\face^t$:
\begin{equation}\label{eq:nn_tri}
	\mathcal{A}(\face^t)=\{\face_j|\face_j\cap\face^t\neq \phi, \face_j\in F\},
\end{equation}
where each adjacent triangle $\face_j$ should share at least one vertex with the current bound triangle $\face^t$. This significantly reduces the search scope from hundreds of thousands triangles to $\sim$12.

Finally, the updated triangle $\face^{t+1}$ is defined as the adjacent triangle that is nearest to the Gaussian center $\x^{t+1}$:
\begin{equation}
\face^{t+1}=\argmin_{\face_j}\min_{\mathbf{p}\in \face_j}{\|\x^{t+1}-\mathbf{p}\|^2},\quad \face_j\in\mathcal{A}(\face^t).
\end{equation}
The whole process is summarized in Algorithm~\ref{alg:walk_on_mesh}.

\begin{algorithm}
\caption{Walking on mesh}\label{alg:walk_on_mesh}
\begin{algorithmic}[1]
\State \textbf{Input: $\x^t,\face^t, \x^{t+1}, F$}
\State \textbf{Output: $\face^{t+1}$}
\State $\bary\gets ComputeBaryCoordinates(\x^{t+1}, \face^t)$\Comment{Eq.~\ref{eq:proj_bary}}
\If{OutOfTrianlge($\bary$)}\Comment{Eq.~\ref{eq:oot}}
	\State $\mathcal{A} \gets AdjacentFaceSet(\face^t, F)$\Comment{Eq.~\ref{eq:nn_tri}}
	\For{Faces $\face_j\in \mathcal{A}$ } 
		\State $D[j]\gets PointMeshDistance(\x^{t+1},\face_j)$
	\EndFor
	\State $j\gets \argmin_{j} D$
	\State \Return $\face_j$	
\Else
	\State \Return $\face^t$
\EndIf
\end{algorithmic}
\end{algorithm}

\subsection{Pose-driven Gaussian Deformation \& Colorization}\label{sec:def_color}

We use the pose-dependent deformation and colorization modules to warp the canonical Gaussians to the observation space and compensate the appearance changes. It includes an articulated deformation module, a non-rigid deformation module, and a pose-dependent colorization module.

\noindent
\textbf{Articulated deformation module.}
To render Gaussians under arbitrary human poses $\pose$, we use a forward articulated deformation function $\mathcal{W}$ to warp the canonical Gaussians $\mathcal{G}_c$ to the posed Gaussians $\gaus_o$ in the observation space:
\begin{equation}
\gaus_o=\mathcal{W}(\mathcal{G}_c, \pose),
\end{equation}
Specifically, we apply Linear Blend Skinning (LBS) as the warp function $\mathcal{W}$, which defines a local transformation $[\mathbf{A}, \mathbf{b}]$ for each Gaussian based on the input human pose $\pose$:
\begin{align}\label{eq:db}
	[\mathbf{A}, \mathbf{b}]=
	\sum_{b=1}^{B}w_b(\x_c) [\mathbf{R}_b(\pose),\mathbf{t}_b(\pose)],
\end{align}
where ${\bf x_c}\in \mathbb{R}^3$ is the Gaussian center in the canonical space,  $b\in{1,...,B}$ denotes $B$ human bones, $\mathbf{R}_b(\pose)\in\operatorname{SO}(3)$ and $\mathbf{t}_b(\pose)\in\mathbb{R}^3$ is the rotation and translation of bone $b$ depending on the human pose $\pose$, $[\cdot,\cdot]$ denotes the concatenation, and $w_b$ is the blending weight of the point $\x_c$. Finally, the outputs are the blended rotation $\mathbf{A}\in\mathbb{R}^{3\times 3}$, and translation $\mathbf{b}\in\mathbb{R}^{3}$.

Then, the canonical Gaussians are transformed to the observation space by:
\begin{equation}
	\x_o =\mathbf{A}\x_c+\mathbf{b}, \quad \mathbf{\gcov}_o=\mathbf{A}\mathbf{\gcov}_c\mathbf{A}^T.
\end{equation}

\noindent
\textbf{Non-rigid Gaussian deformation.}
The above articulated deformation alone cannot capture fine-grained clothes deformation. Thus, we further introduce a non-rigid deformation module~\cite{peng2024animatable, weng2022humannerf, jiang2022neuman,qian20233dgs,yang2023deformable}. 

To fit the high-frequency non-rigid deformation efficiently, we adopt the multi-resolution hash encoder $\operatorname{MHE}$~\cite{instant-ngp} as the spatial encoder, and decode the deformation via a shallow MLP:
\begin{equation}
	\Delta{\x}, \Delta \mathbf{s}, \Delta \quater = \operatorname{MLP_{NR}}(\operatorname{MHE}(\x), f(\pose)),
\end{equation}
where the input is the spatial embedding $\operatorname{MHE}(\x)$ and a latent code $f(\pose)$, which encodes human pose $\pose$ via a human pose encoder $f$~\cite{mihajlovic2021leap}, and the outputs are the center offset $\Delta\x$, rotation offset $\Delta \quater$ represented as quaternions and the scale changes $\Delta \mathbf{s}$.

Since the non-rigidly deformed Gaussians will deviate from the mesh, we only apply the module in the second Detached Stage.

\paragraph{Pose-driven colorization.}
Human motion causes shadows on the surface. This leads to inconsistent textures for the same Gaussians rendered in different frames. Due to the strong expressive ability of 3D Gaussians, directly learning appearance by optimizing per-Gaussian color leads to overfitting. Thus, we use an MLP to learn pose-dependent Gaussian color $\mathbf{c}$. Specifically, we condition the color with a per-frame latent vector $\boldsymbol{\psi}\in\mathbb{R}^{16}$ for global lighting change due to self-rotating, and a pose-aware latent vector $h(\pose)$ for local shading caused by wrinkles:
\begin{equation}
\mathbf{c}=\operatorname{MLP_{RGB}}(\x,\boldsymbol{\psi},h(\boldsymbol{\theta})),
\end{equation}
where we obtain $h(\boldsymbol{\theta})$ by extracting the latent feature from the intermediate layer of the non-rigid network $\operatorname{MLP_{NR}}$.

\section{Optimization}
\subsection{Training objective}
The final training objective function is composed of three main terms:
\begin{equation}\label{eq:obj_func}
	L=L_{\rm app}+ L_{\rm geo} + L_{\rm smooth},
\end{equation}
where $L_{\rm geo}$ is the Gaussian-Mesh alignment term introduced in Sec.~\ref{sec:reg}, and $L_{\rm smooth}$ is the smooth regularization defined in Eq.~\ref{eq:smooth}.

For $L_{\rm app}$, we minimize the difference between the rendered and ground-truth images, defined as:
\begin{equation}
L_{\rm app}=L_1+\lambda_{\rm mask}L_{\rm mask}+\lambda_{\rm LPIPS} L_{\rm LPIPS},
\end{equation}
where the $L_1$ and $L_{\rm mask}$ minimize the L1 norm of the RGB difference and opacity difference between the rendered and ground-truth images, respectively. $L_{\rm LPIPS}$ is the perceptual loss~\cite{zhang2018perceptual}, which improves image sharpness when small misalignment exists.

\subsection{Implementation details}
SAGA is trained for 15k iterations, with 3k iterations for the first Adhered Stage and 12k iterations for the Detached Stage on one NVIDIA RTX3090 GPU, taking 12 minutes on average. We simultaneously optimize the Gaussian parameters, non-rigid deformation module and the colorization module with the Adam optimizer~\cite{adam}. The Gaussian parameters are optimized with the same learning rate and scheduler as used in 3DGS~\cite{gaussplat}. We mute the non-rigid deformation until after 3k iterations. The initial learning rate of the non-rigid and colorization networks is set as $1\times 10^{-3}$ and gradually decayed by $0.1$ with the exponential scheduler.
During inference, SAGA renders a 512$\times$512 image in real-time at 60 FPS on one NVIDIA RTX3090 GPU. Please refer to the supplementary material for more details.

\section{Experiment}
We evaluate SAGA against state-of-the-art methods for novel view and pose synthesis, as well as geometry reconstruction, using monocular videos from challenging datasets~\cite{alldieck2018video, peng2021neural, peng2024animatable}.

\begin{figure*}[htbp]
	\centering
	\includegraphics[width=1.0\linewidth]{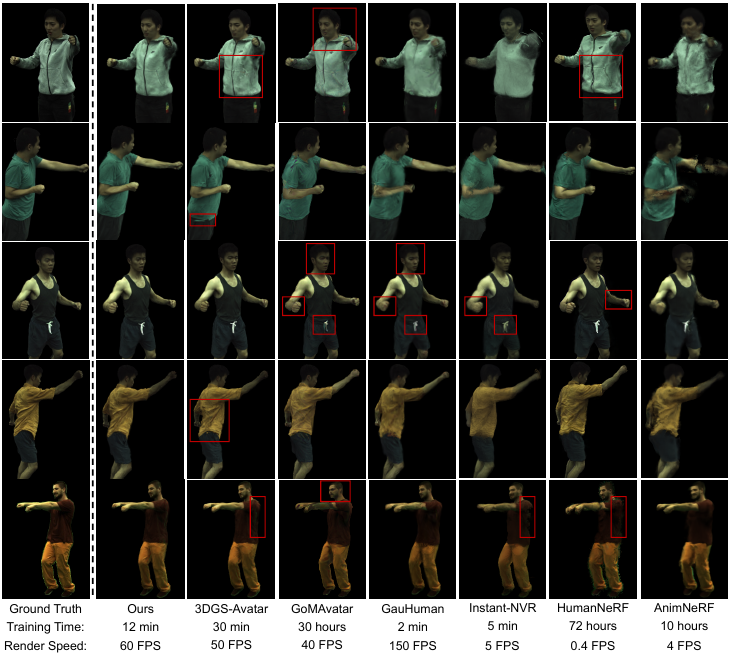}
	\caption{\textbf{Comparison results of novel view synthesis on ZJU-MoCap dataset~\cite{peng2021neural} and MonoCap dataset~\cite{peng2024animatable}.} For Gaussian based methods, 3DGS-Avatar~\cite{qian20233dgs} suffers from artifacts, GoMAvatar~\cite{gomavatar} struggles to fit details in the faces and hands, and GauHuman~\cite{gauhuman} synthesizes oversmoothed results, while other NeRF-based methods generally suffer from blurred~\cite{instant-nvr, peng2021animatable} or distorted results~\cite{weng2022humannerf}. In contrast, our method achieves more photorealistic rendering results, and is also more efficient than most methods except for GauHuman and Instant-NVR, which, however, cannot fit high-frequency details within such short training time.}
	\label{fig:zju}
\end{figure*}
\subsection{Datasets}

\noindent
{\bf ZJU-MoCap}~\cite{peng2021neural} dataset contains 23-view videos of 9 human subjects performing complex dynamic motions, such as kicking and swirling. We conduct novel view synthesis on 6 commonly used subjects~\cite{weng2022humannerf}. For training, we select $\sim$500 frames from \emph{camera 4} for training, and use the rest 22 views for evaluation.

\noindent
{\bf MonoCap} dataset~\cite{peng2024animatable} consists of multi-view videos from DeepCap dataset~\cite{habermann2020deepcap} and DynaCap dataset~\cite{habermann2021real} selected by~\cite{peng2024animatable}. We use 500 frames from one view for training and evaluate on 10 other novel views that distribute uniformly, following~\cite{instant-nvr,gauhuman}.

\noindent
{\bf PeopleSnapshot}~\cite{alldieck2018video} dataset contains monocular videos of humans that self-rotate in a fixed A-pose. We conduct experiments on 4 subjects, and follow the protocol of Anim-NeRF~\cite{chen2021animatable} to train all comparison methods on their optimized pose parameters. We evaluate on images rendered under a white background following~\cite{chen2021animatable, instantavatar}.

\subsection{Baselines}
We compare our proposed SAGA with \textbf{NeRF-based methods}~\cite{peng2021neural, peng2021animatable, weng2022humannerf, instant-nvr, instantavatar}, and more recent \textbf{3DGS-based methods}~\cite{qian20233dgs,gomavatar,gauhuman,splattingavatar}.
For NeRF-based methods:
NeuralBody~\cite{peng2021neural} anchors latent codes on the SMPL mesh and diffuse in 3D space with 3D convolution networks for neural volume rendering;
AnimatableNeRF~\cite{peng2021animatable} represents the scene with a large MLP, and learns a forward and backward blending weight MLP for animation;
HumanNeRF~\cite{weng2022humannerf} further incorporates a non-rigid deformation network and achieves SOTA performance;
InstantAvatar~\cite{instantavatar} applies the efficient Instant-NGP~\cite{instant-ngp} as the canonical representation;
InstantNVR~\cite{instant-nvr} designs multi-part hash encoder based on~\cite{instant-ngp}.
For Gaussian based methods, GauHuman~\cite{gauhuman} applies naive Gaussians in the canonical space, and optimizes a blending weight offset network;
3DGSAvatar~\cite{qian20233dgs} further learns the non-rigid deformation and the pose-dependent colors;
GoMAvatar~\cite{gomavatar} binds the Gaussians rigidly on the mesh with fixed location, and does not align the orientation;
SpattingAvatar~\cite{splattingavatar} parameterizes Gaussian location with uv coordinates and offset $d$ along the mesh normal, but fixes the mesh vertices.
\definecolor{tabfirst}{rgb}{1, 0.7, 0.7}
\definecolor{tabsecond}{rgb}{1, 0.85, 0.7}
\definecolor{tabthird}{rgb}{1, 1, 0.7}
\begin{table*}[htbp]
	\centering
	\setlength{\tabcolsep}{1.pt}
	\resizebox{\textwidth}{!}{
		\begin{tabular}{l|c|cc|ccc|ccc|ccc|ccc|ccc|ccc}
			\toprule
			\multicolumn{2}{c|}{Subjects}&&&\multicolumn{3}{c|}{\textbf{377}} & \multicolumn{3}{c|}{\textbf{386}} & \multicolumn{3}{c|}{\textbf{387}} & \multicolumn{3}{c|}{\textbf{392}} & \multicolumn{3}{c|}{\textbf{393}} & \multicolumn{3}{c}{\textbf{394}}   \\
			\midrule
			\multicolumn{2}{c|}{Metrics} &Train$\downarrow$& FPS$\uparrow$ & PSNR$\uparrow$ & SSIM$\uparrow$ & LPIPS$\downarrow$ & PSNR$\uparrow$ & SSIM$\uparrow$ & LPIPS$\downarrow$ & PSNR$\uparrow$ & SSIM$\uparrow$ & LPIPS$\downarrow$ & PSNR$\uparrow$ & SSIM$\uparrow$ & LPIPS$\downarrow$ & PSNR$\uparrow$ & SSIM$\uparrow$ & LPIPS$\downarrow$ & PSNR$\uparrow$ & SSIM$\uparrow$ & LPIPS$\downarrow$  \\
			\midrule
			\multirow{4}{*}{\rotatebox{90}{NeRF}}
			&NeuralBody~\cite{peng2021neural} &10h  &  4
			& 29.11 & 0.9674 & 40.95 
			& 30.54 & 0.9678 & 46.43 
			& 27.00 & 0.9518 & 59.47 
			& 30.10 & 0.9642 & 53.27 
			& 28.61 & 0.9590 & 59.05 
			& 29.10 & 0.9593 & 54.55 \\
			&AnimNeRF~\cite{peng2021animatable} &	10h  &  4 
			&29.12& 0.9727& 26.58& 32.94& 0.9695& 36.04& 27.93& 0.9601& 41.76& 29.50& 0.9635& 39.45& 27.64& 0.9566 &43.17 &29.15& 0.9595& 38.08\\
			
			&HumanNerf~\cite{weng2022humannerf}  & 72h & 0.4
			& 31.12 & 0.9774 & 22.80 
			& 33.31 & 0.9726 & 33.48 
			& \textbf{28.27} & 0.9617 & 38.89
			& 31.34 & 0.9712 & 33.57
			& 29.19 & 0.9644 & 36.88
			& 30.74 & 0.9662 & 34.67  \\
			&InstantNVR~\cite{instant-nvr}	&	\underline{0.1h}	&	5
			& 31.36 & 0.979 & 26.03
			& 33.53 & 0.977 & 33.02
			& 28.11 & 0.963 & 46.96
			& 32.03 & 0.973 & 39.30
			& 29.55 & 0.964 & 46.29
			& 31.39 & 0.969 & 40.00 \\
			\multirow{4}{*}{\rotatebox{90}{Gaussian}}
			&GoMAvatar~\cite{gomavatar} &	30h	&	40
			&  31.11 &  0.9787 &  21.04
			&  33.26 &  0.9764 &  26.63
			&  27.87 & 	0.9616 &  37.03 
			&  31.28 & 	0.9721 &  30.75 
			&  29.06 &  0.9640 &  33.97 
			&  30.46 &  0.9655 &  31.12  
			\\
			&GauHuman~\cite{gauhuman}	&	\textbf{2min} & \textbf{150}
			& 32.04 & 0.9751 & 19.13
			& 33.77	& 0.9681 & 28.62
			& 28.26	& 0.9548 & 38.72
			& 32.17 & 0.9648 & 30.02
			& 29.75 & 0.9565 & 35.27
			& 31.46 & 0.9589 & 30.82 \\
			
			&3DGSAvatar~\cite{qian20233dgs} &	0.5h & 50 
			& 30.88  & 0.9785  &  19.09
			& 33.38 & 0.9772 &  25.65
			& 27.75  & 0.9630	&  34.71
			& 31.88  & 0.9742	&  29.30
			& 29.28  & 0.9659 &  32.46
			& 30.68  & 0.9679 & 28.93 \\
			&\textbf{Ours}   &	0.2h	&	\underline{60}
			& \textbf{31.64} & \textbf{0.9806} & \textbf{18.61}
			& \textbf{33.80} & \textbf{0.9788} & \textbf{25.54}
			& 28.06 & \textbf{0.9648} & \textbf{34.20}
			& \textbf{32.30} & \textbf{0.9752} & \textbf{28.83}
			& \textbf{29.77} & \textbf{0.9680} & \textbf{32.15}
			& \textbf{31.29} & \textbf{0.9699} & \textbf{28.48} \\
			\bottomrule 
		\end{tabular}  
	}\caption{
	\textbf{Qualitative results of novel view synthesis on ZJU-MoCap~\cite{peng2021neural}.} Performance is evaluated with PSNR, SSIM and LPIPS metrics. Our method outperforms all comparison methods on SSIM and LPIPS, while being more efficient than most of them in training and rendering. Though InstantNVR and GauHuman are faster, our method can surpass them on challenging subjects by $37\%$ and $13\%$.
	}\label{tab:zju_novel_view}
\end{table*}
\begin{table*}[htbp]
	\resizebox{\textwidth}{!}{
		\begin{tabular}{c|r|c|ccc|ccc|ccc|ccc}
			\toprule
			\multicolumn{2}{c|}{Subjects}
			&& \multicolumn{3}{c|}{\textbf{Lan}} & \multicolumn{3}{c|}{\textbf{Marc}} & \multicolumn{3}{c|}{\textbf{Olek}} & \multicolumn{3}{c}{\textbf{Vlad}}\\
			\midrule
			\multicolumn{2}{c|}{Metrics}
			&Train$\downarrow$& PSNR$\uparrow$  & SSIM$\uparrow$  & LPIPS$\downarrow$  & PSNR$\uparrow$  & SSIM$\uparrow$  & LPIPS$\downarrow$  & PSNR$\uparrow$  & SSIM$\uparrow$  & LPIPS$\downarrow$  & PSNR$\uparrow$  & SSIM$\uparrow$  & LPIPS$\downarrow$ \\
			\midrule
			\multirow{3}{*}{\rotatebox{90}{NeRF}}&
			AnimNeRF~\cite{peng2024animatable}& 10 hours
			&	31.40	&	0.9863	& 	0.0183 
			&	30.81	&	0.9834	& 	0.0242
			&	34.18	&	0.9880	& 	0.0155 
			&	27.90	&	0.9810	& 	0.0200  \\
			&HumanNeRF~\cite{weng2022humannerf}& 72 hours
			& 33.50 & 0.9895 & 0.0134	
			& 34.66 & 0.9904 & 0.0164 
			& 34.08 & 0.9895 & 0.0143 
			& 28.49 & 0.9814 & 0.0178  \\
			&InstantNVR~\cite{instant-nvr}& \underline{10 min}
			& 32.78 & 0.9871 & 0.0171
			& 33.84 & 0.9894 & 0.0169  
			& 34.52 & 0.9892 & 0.0139 
			& 28.70 & 0.9830 & 0.0197  \\
			\midrule
			\multirow{4}{*}{\rotatebox{90}{Gaussian}}
			&GauHuman~\cite{gauhuman} & \textbf{2 min}
			& 33.53 & 0.9852 & 0.0108
			& 34.68 & 0.9855 & 0.0159
			& \underline{34.77} & 0.9850 & 0.0134
			& 28.46 & 0.9796 & 0.0182	\\
			&GoMAvatar~\cite{gomavatar} & 30 hours
			& 33.37 & 0.9887 & 0.0120
			& 34.17	& 0.9881 & \underline{0.0128}
			& 34.41 & 0.9866 & 0.0134
			& 27.55 & 0.9782 & 0.0202 \\
			&3DGSAvatar~\cite{qian20233dgs}& 30 min
			& \underline{34.27} & \underline{0.9900} & \underline{0.0099}
			& \underline{35.40} & \underline{0.9906} & 0.0134
			& 34.69 & \underline{0.9893} & \textbf{0.0115}
			& \textbf{29.06} & \underline{0.9837} & \textbf{0.0158}	\\
			&\textbf{Ours}& 12 min
			& \textbf{34.37} & \textbf{0.9901} & \textbf{0.0098} 
			& \textbf{35.59} & \textbf{0.9908} & \textbf{0.0128} 
			& \textbf{34.86} & \textbf{0.9895} & \underline{0.0116}
			& \underline{28.92} & \textbf{0.9837} & \underline{0.0165}\\
			\bottomrule
		\end{tabular}
	}\caption{\textbf{Quantitative results of novel view synthesis on MonoCap dataset~\cite{peng2024animatable}}. Our method outperforms the comparison methods on most subjects with fast training speed. }\label{tab:monocap}
\end{table*}

\subsection{Comparison Results on Novel View Synthesis}

\subsubsection{Results on ZJU-MoCap and Monocap dataset}
We evaluate all comparison methods by running their official code. Differences from the results reported in the original papers are due to different settings including the training view, frame number, or implementation of evaluation metrics, which are all unified in our experiments for a fair comparison.

We present quantitative results in Tab.~\ref{tab:zju_novel_view}. 
Our proposed SAGA consistently outperforms the state-of-the-art Gaussian and NeRF-based methods on the SSIM and LPIPS metrics. Moreover, SAGA also achieves \emph{third} highest training efficiency of $\sim$12 minutes and \emph{second} highest real-time rendering speed at 60 FPS, which is comparable to the most efficient methods InstantNVR~\cite{instant-nvr} and GauHuman~\cite{gauhuman}. Although they can be trained more efficiently, they struggle at fitting fine deformation within such short training time. As a result, our method can notably surpass them on challenging subject by $37\%$ and $13\%$ respectively. Notably, compared to the only method that aligns Gaussian on mesh, i.e., GoMAvatar~\cite{gomavatar}, our method is 150$\times$ faster while achieving higher rendering quality. We attribute this to the higher flexibility of our two-stage Gaussian alignment strategy over their rigidly binding on mesh strategy. Since their overconstrained Gaussians struggle to move to the real surface, it requires more steps of careful optimization to converge.

For the qualitative results shown in Fig.~\ref{fig:zju}, our method can synthesize higher quality results with more realistic details, such as clothes wrinkles, fists and human faces, while former NeRF-based method typically generates oversmoothed results. HumanNeRF~\cite{weng2022humannerf} performs well but requires three days of training and still produces artifacts in thin structures like zippers. For Gaussian based methods, 3DGS-Avatar~\cite{qian20233dgs} suffers from noisy textures in novel views (Row 1), and fails to synthesize consistent results at details, such as the zipper (Row 1) and the lower rim of the shirt (Row 2). GoMAvatar~\cite{gomavatar} struggles with facial realism and detailed features like fists and drawstrings due to its fixed-on-mesh representation. In contrast, our method can synthesize more plausible and photorealistic results with well-preserved details, demonstrating the superiority of the proposed surface-aligned representation in regulating Gaussians to improve multi-view consistency without compromising  the rendering realism.

\subsubsection{Results on PeopleSnapshot dataset}
\begin{figure*}[htbp]
	\centering
	\includegraphics[width=1.0\linewidth]{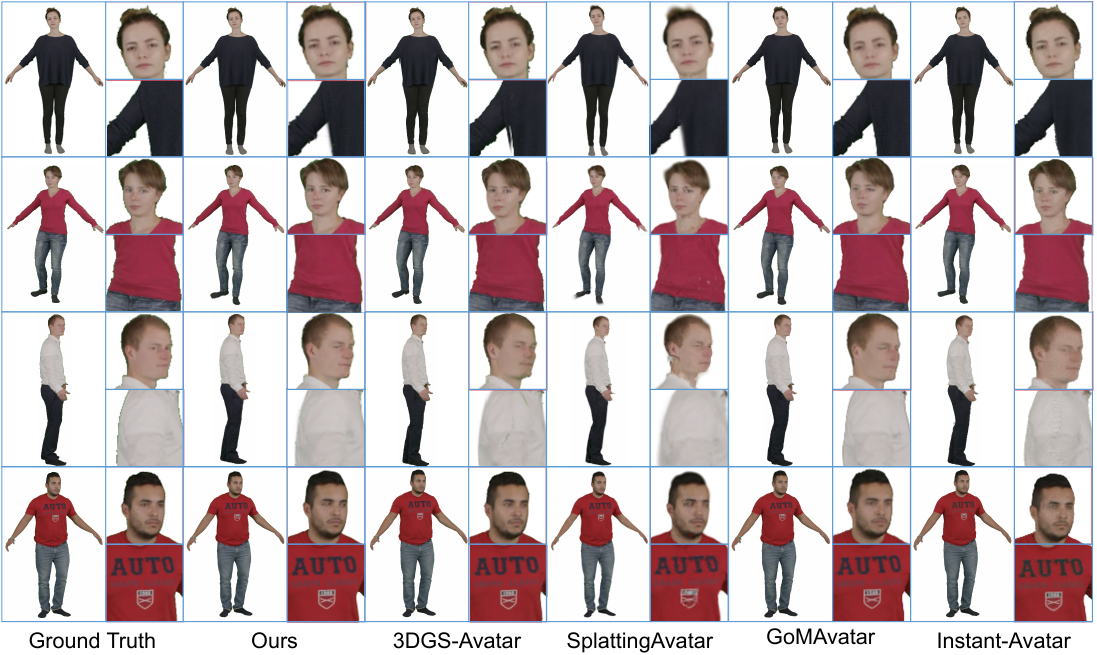}
	\caption{
		\textbf{Comparison results of novel view synthesis with State-of-the-art methods on PeopleSnapshot Dataset~\cite{alldieck2018video}.} All methods are Gaussian-based, except the last column that is NeRF-based. Our method can synthesize more photorealistic images, which are free of the artifacts (Row 1 Col. 3, Row 2 Col. 4, Row 3 Col. 4) and distortions (Row 2 Col. 3) in 3DGS-Avatar~\cite{qian20233dgs} and SplattingAvatar~\cite{splattingavatar}, while being shaper than GoMAvatar~\cite{gomavatar} and InstantAvatar~\cite{instantavatar} with higher fidelity.}
	\label{fig:peoplesnapshot}
\end{figure*}

As shown in Tab.~\ref{tab:peoplesnapshot}, our method surpasses the comparison methods on all subjects in terms of LPIPS metric. Though SplattingAvatar~\cite{splattingavatar} achieves higher PSNR and SSIM metrics on the \textit{Female-3-causal} subject, we believe it is because these metrics favor the smoothed blurry results over sharper but slightly misaligned grid-like texture of the sweater (Fig.~\ref{fig:peoplesnapshot} Col. 4). Contrastively, our method preserves the details of the sweater, and surpasses SplattingAvatar by $42\%$ in terms of LPIPS, which we believe to be more persuasive under this scene.

We show qualitative results in Fig.~\ref{fig:peoplesnapshot}. Our method synthesizes more photorealistic results, especially for finer structures such as hands, faces, and wrinkles.
NeRF-based method~\cite{instantavatar} suffers from salt noises from the close-up images due to the ray sampling in the volume rendering process. For Gaussian-based methods, without proper regularization, 3DGS-Avatar and SplattingAvatar tend to generate artifacts at the armpit and shoulder, while mesh-based GoMAvatar struggles to synthesize realistic human faces (Row 2).

\begin{table*}[htbp]
	\resizebox{\textwidth}{!}{
	\begin{tabular}{c|r|c|ccc|ccc|ccc|ccc}
		\toprule
		\multicolumn{2}{c|}{Subjects}
		&& \multicolumn{3}{c|}{\textbf{male-3-casual}} & \multicolumn{3}{c|}{\textbf{male-4-casual}} & \multicolumn{3}{c|}{\textbf{female-3-casual}} & \multicolumn{3}{c}{\textbf{female-4-casual}}\\
		\midrule
		\multicolumn{2}{c|}{Metrics}
		&Train$\downarrow$& PSNR$\uparrow$  & SSIM$\uparrow$  & LPIPS$\downarrow$  & PSNR$\uparrow$  & SSIM$\uparrow$  & LPIPS$\downarrow$  & PSNR$\uparrow$  & SSIM$\uparrow$  & LPIPS$\downarrow$  & PSNR$\uparrow$  & SSIM$\uparrow$  & LPIPS$\downarrow$ \\
		\midrule
		\multirow{3}{*}{\rotatebox{90}{NeRF}}
		&Neural Body~\cite{peng2021neural}& $\sim$ 14h 
		& 24.94 & 0.9428 & 0.0326 & 24.71 & 0.9469 & 0.0423& 23.87& 0.9504& 0.0346& 24.37& 0.9451 & 0.0382 \\
		&Anim-NeRF~\cite{chen2021animatable}& $\sim$ 13h 
		&29.37 & 0.9703 & 0.0168 
		& 28.37 & 0.9605 & \underline{0.0268} 
		& 28.91 & 0.9743 & 0.0215 
		&28.90 & 0.9678 & 0.0174 \\
		&InstantAvatar~\cite{instantavatar}& \textbf{1 min}
		& 29.65 & 0.9730 & 0.0192 & 27.97 & 0.9649 & 0.0346  & 27.90 & 0.9722 & 0.0249 & 28.92 & 0.9692 & 0.0180  \\
		\midrule
		\multirow{4}{*}{\rotatebox{90}{Gaussian}}
		&3DGSAvatar~\cite{qian20233dgs}& 45 min
		& 31.82 & \underline{0.9800} & 0.0196
		& 29.67 & 0.9755 & 0.0302
		& 29.49 & 0.9736 & 0.0214
		& 30.28 & 0.9769 & 0.0188	\\
		&SplattingAvatar~\cite{splattingavatar}	& 18 min
		& \underline{32.47}	& 0.9784 & 0.0243
		& \underline{30.74}	& \underline{0.9764} & 0.0347
		& \textbf{30.65}	& \textbf{0.9784} & 0.0343
		& 31.19	& 0.9763 & 0.0287	\\
		&GoMAvatar~\cite{gomavatar} & 20 h
		& 31.74 & 0.9793 & \underline{0.0187}
		& 29.78 & 0.9738 & 0.0282
		& 29.83 & 0.9758 & \underline{0.0209}
		& \underline{31.38} & \underline{0.9780} & \underline{0.0174}	\\
		&\textbf{Ours}& \underline{12 min}
		& \textbf{33.15} & \textbf{0.9828} & \textbf{0.0135} 
		& \textbf{30.95} & \textbf{0.9776} & \textbf{0.0248} 
		& \underline{29.85} & \underline{0.9760} & \textbf{0.0196}
		& \textbf{31.92} & \textbf{0.9784} & \textbf{0.0145}\\
		\bottomrule
	\end{tabular}
}\caption{ \textbf{Quantitative results on the PeopleSnapshot dataset~\cite{alldieck2018video}.} Our method outperforms all the Gaussian and NeRF-based methods on LPIPS metric while being the second fastest method in terms of training time, and notably surpasses the fastest method InstantAvatar~\cite{instantavatar} in terms of rendering quality.}
\label{tab:peoplesnapshot}
\end{table*}

\subsection{Comparison results on novel pose synthesis}
To evaluate novel pose synthesis performance, we animate the reconstructed Gaussians with
out-of-distribution pose sequences from AIST++~\cite{aist} and AMASS~\cite{amass} datasets, containing complex dancing motions.

\subsubsection{Novel pose synthesis results on ZJU-MoCap dataset}

\begin{figure*}[htbp]
	\begin{center}
		\includegraphics[width=0.90\linewidth]{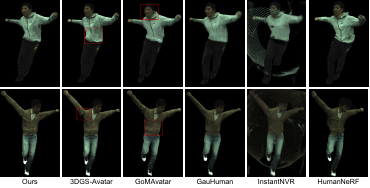}
		\caption{\textbf{Comparison results of animating ZJU-MoCap subjects with out-of-distribution poses from AIST++ dataset~\cite{aist}.} Our method can synthesize more natural results without undesired artifacts.}
		\label{fig:zju_pose}
	\end{center}
\end{figure*}

As shown in Fig.~\ref{fig:zju_pose}, our method produces consistently high-quality results on out-of-distribution poses with well-preserved details at the zipper, hands and even tiny buttons. In contrast, the comparison methods generally suffer from artifacts. For instance, 3DGS-Avatar~\cite{qian20233dgs} synthesizes blurred zipper, and fractures at the armpit. GoMAvatar~\cite{gomavatar} and GauHuman~\cite{gauhuman} are limited by lower-quality reconstructions during training. Specifically, GoMAvatar synthesizes distorted faces and hands due to the constrained fitting ability of its fixed-on-mesh representation. GauHuman produces oversmoothed results. For NeRF-based methods, Instant-NVR~\cite{instant-nvr} suffers from severe artifacts due to the poor generalization ability of their multi-part hash encoder when applied to novel poses. HumanNeRF~\cite{weng2022humannerf} produces unnatural deformations leading to distorted faces and bodies.

\subsubsection{Novel pose results on PeopleSnapshot dataset}
Animating the subjects from the PeopleSnapshot dataset with novel poses can be more challenging because the pose variety in this dataset is more limited, containing only self-rotating A-pose.

\begin{figure*}[htbp]
	\begin{center}
		\includegraphics[width=1.0\linewidth]{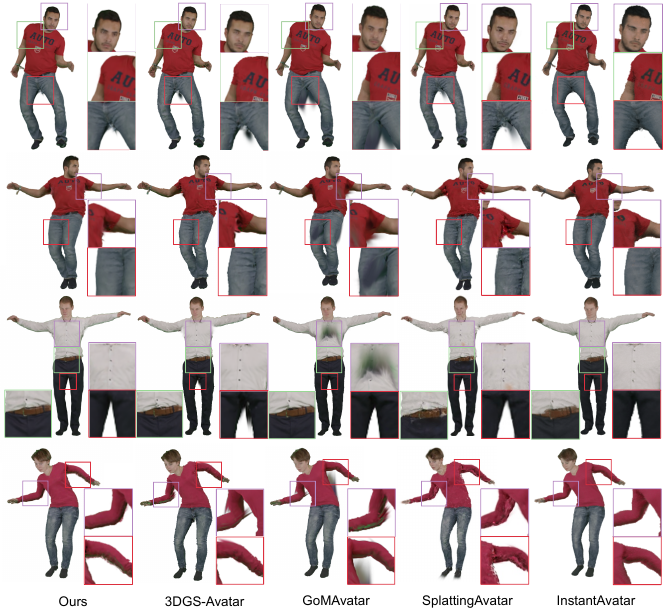}
		\caption{\textbf{Comparison results of animating PeopleSnapshot subjects with out-of-distribution poses from AIST++ dataset~\cite{aist}.}. Our method can synthesize more photorealistic and plausible results, while 3DGS-Avatar~\cite{qian20233dgs} and GoMAvatar~\cite{gomavatar} suffer from artifacts.}
		\label{fig:ppss_pose}
	\end{center}
\end{figure*}

As shown in Fig.~\ref{fig:ppss_pose}, our synthesized images maintain the high quality as on the seen poses. Contrastively, 3DGS-Avatar~\cite{qian20233dgs} suffers from needle-like artifacts at the joints, due to the inaccurate LBS weights learned during training. Our surface-aligned representation allows the Gaussians to inherent more natural LBS weight from the SMPL mesh, effectively avoiding such issues. Moreover, GoMAvatar~\cite{gomavatar} produces artifacts stemming from oversized Gaussians. 
This is likely because GoMAvatar only aligns Gaussian center on the mesh, but ignores the orientation, leading to overfitting to the low-variety A-pose. Consequently, these oversized Gaussians produce artifacts when animated with novel poses.
In contrast, the proposed SAGA alleviates this problem by aligning both position and orientation, demonstrating the superior generalization ability in out-of-distribution poses.

\subsection{Ablation Study}
In this section, we discuss the effect of various strategies proposed in SAGA including the \emph{Two-stage Surface-Aligned Gaussian representation}, the \emph{Gaussian-Mesh Alignment losses}, and the \emph{Walking-on-Mesh strategy}.

\begin{figure*}[htbp]
	\begin{center}
		\includegraphics[width=0.99\linewidth]{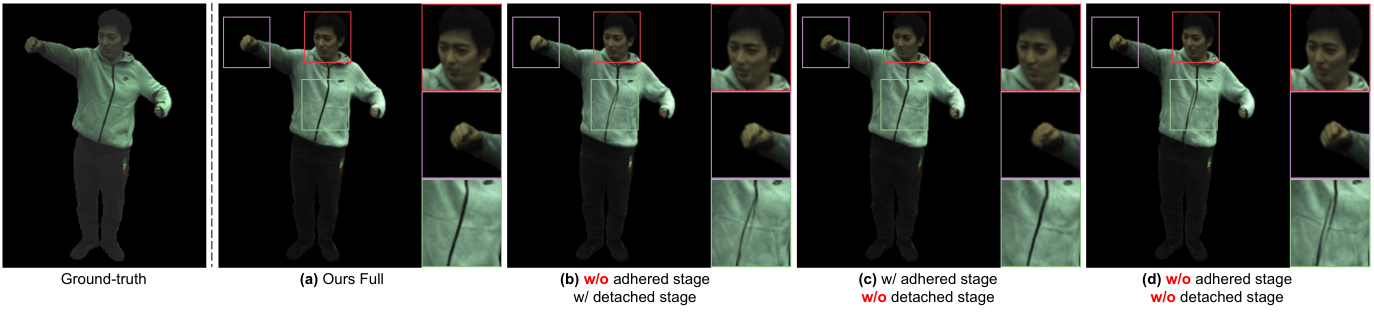}
		\caption{\textbf{Novel view synthesis results of ablation study on the training stages}. \emph{\textbf{Ours Full}} with both the \emph{Adhered }and the \emph{Detached Stage} achieves the most photorealistic results, which avoid artifacts while having sharper details and more natural human expressions.}
		\label{fig:ablation_stage}
	\end{center}
\end{figure*}

\subsubsection{Effect of the two-stage Surface-Aligned Gaussians}
We evaluate the individual effects of the first \emph{Adhered Stage} and the second \emph{Detached Stage} by removing each of them respectively. The mean results across 6 instances on ZJU-MoCap dataset are reported in Tab.~\ref{tab:ablation_zju_stage}. Our full two-stage SAGA representation achieves the best performance. We show the qualitative results in Fig.~\ref{fig:ablation_stage}, and analyze them as follows.

\noindent
\textbf{Effect of the Adhered Stage.}
The Adhered Stage constrains the Gaussians to stay strictly on the surface, which we expect to guide the Gaussians to learn well-defined geometry in the early training stage. As shown in Fig.~\ref{fig:ablation_stage}(d), without this stage, artifacts such as diverged zippers emerge. This shows that directly optimizing Gaussians cannot ensure the multi-view consistency of such thin structure. For \textbf{w/o adhered stage, w/ detached stage} shown in Fig.~\ref{fig:ablation_stage}(b),
though the Gaussian-Mesh alignment regularization in the Detached Stage can mitigate the issue of diverged zipper, it is not powerful enough to regularize the Gaussians from scratch, compared to the strict alignment of the Adhered Stage .

\noindent
\textbf{Effect of the Detached Stage.} The Detached Stage further unleashes the fitting potential of the Gaussians from the Adhered Stage by allowing them to move freely near the surface during optimization. Fig.~\ref{fig:ablation_stage} (c) shows that omitting  this stage results in blurrier details at the fists and the unnatural face expressions. Ultimately, our full two-stage SAGA renders sharper images without any artifacts.

\begin{figure}[htbp]
	\begin{center}
		\includegraphics[width=0.98\linewidth]{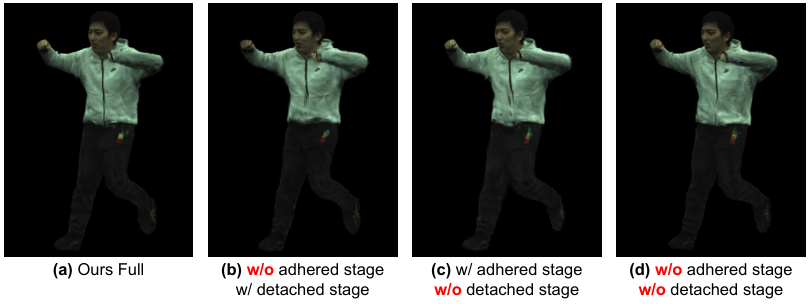}
		\caption{\textbf{Novel pose synthesis results of ablation study on the training stages}, ours full synthesizes more photorealistic results, while the others suffer from blurry details at the zipper.}
		\label{fig:zju_ablation_stage_novel_pose}
	\end{center}
\end{figure}

\noindent
\textbf{Effect on novel pose synthesis.}
As shown in Fig.~\ref{fig:zju_ablation_stage_novel_pose}, our full two-stage representation generates more plausible results, indicating that our SAGA learns canonical Gaussians with improved geometry, leading to better generalization on novel poses.

\begin{table}[htbp]
	\centering
		\resizebox{0.85\linewidth}{!}
	{
		\begin{tabular}{cc|ccc}
			\toprule
			Stage 1 & Stage 2 &  &  & \\
			Adhered & Detached & PSNR$\uparrow$ & SSIM$\uparrow$ & LPIPS$\downarrow$\\
			\midrule
			$\times$ & \checkmark  & 31.05 & 0.9725 & 0.0281 \\
			\checkmark & $\times$ & 31.06 & 0.9726 & 0.0298 \\
			$\times$ & $\times$ & 30.64 & 0.9702 & 0.0290 \\
			\checkmark & \checkmark  &  \textbf{31.11} &  \textbf{0.9728} & \textbf{0.0279}	\\
			\bottomrule
		\end{tabular}
	}
	\caption{\textbf{Ablation study on the two-stage representaion.} Mean results over all subjects from ZJU-MoCap is reported.
	}
	\label{tab:ablation_zju_stage}
\end{table}

\begin{figure}[htbp]
	\begin{center}
		\includegraphics[width=0.99\linewidth]{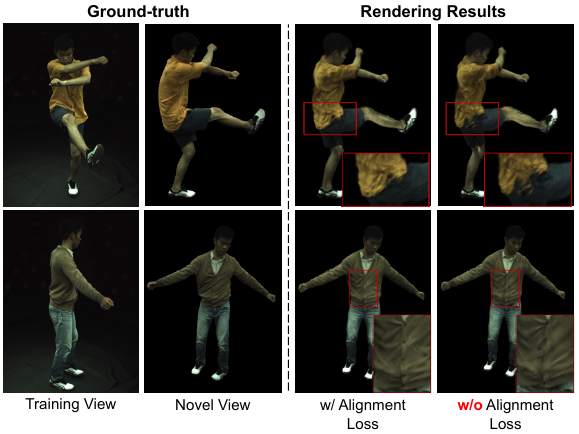}
		\caption{\textbf{Qualitative results of ablation study on the proposed Gaussian-mesh Alignment Regularization.} On the left are the ground-truth images from the training view and novel view, respectively. On the right are the rendering results. Ours with the alignment loss produces more plausible results even in the region barely visible during training.}\label{fig:ablation_zju_alignment_loss}
	\end{center}
\end{figure}
\subsubsection{Effect of the Gaussian-Mesh alignment loss}

The alignment loss regularizes both canonical and deformed Gaussians to align with the underlining surface, and enforces more consistent non-rigid deformation. 
As shown in Fig.~\ref{fig:ablation_zju_alignment_loss}, on the left are the training images from the seen view and the novel view ground-truth images. And on the right are the rendering results. Without the alignment loss, the non-rigid module cannot learn plausible deformation in the area unseen during training, resulting in fractured artifacts (row 1 red box), and misalignment of the buttons (row 2 red box). On the contrary, the proposed alignment loss eliminates the above artifacts, indicating it can effectively regularize the detached Gaussians to learn more natural deformation in the unseen regions, which is also proved by the quantitative results in Tab.~\ref{tab:ablation_zju_loss}.

\begin{table}[htbp]
	\centering
	\resizebox{0.8\linewidth}{!}
	{
		\begin{tabular}{c|ccc}
			\toprule
			& PSNR$\uparrow$ & SSIM$\uparrow$ & LPIPS$\downarrow$\\
			\midrule
			w/o Alignment  & 31.08 & 0.9726 & \textbf{0.0279} \\
			Ours w/ Alignment & \textbf{31.11} & \textbf{0.9728} & \textbf{0.0279} \\
			\bottomrule
		\end{tabular}
	}
	\caption{\textbf{Ablation study on the Alignment Loss.} Mean results over all subjects from ZJU-MoCap is reported.}\label{tab:ablation_zju_loss}
\end{table}

\subsubsection{Effect of the Walking-on-Mesh strategy}\label{sec:ablation_wom}

\begin{figure}[htbp]
	\includegraphics[width=1.0\linewidth]{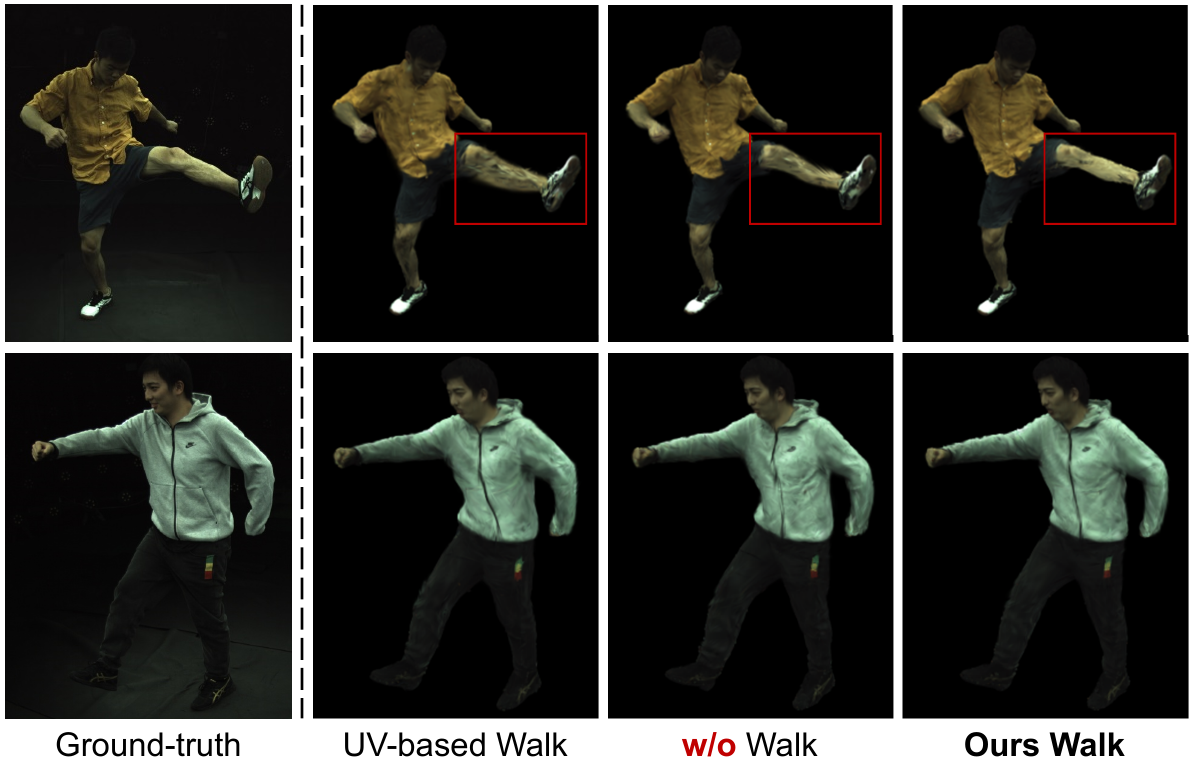}
	\caption{\textbf{Ablation study on the Walking-on-Mesh strategy.} Our method learns more consistent texture at the legs under highly dynamic motion because our Gaussian walking strategy updates the triangles more accurately.}\label{fig:ablation_zju_walking_on_triangles}
\end{figure}

The Walking-on-Mesh Strategy tracks the nearest triangles for the moving Gaussians to ensure them to be aligned with correct triangles. We compare our strategy with the \textbf{UV-based walk} used in SplattingAvatar~\cite{splattingavatar}, which is originally developed by lifted optimization techniques~\cite{shen2020phong, taylor2014user, taylor2016efficient}. It walks in the unfolded triangle space based on the UV updates. 
As shown in Tab.~\ref{tab:ablation_zju_walking_on_triangles}, our method achieves the best performance. We think the reason is that our reprojection based method finds more accurate closest triangles in the Euclidean space. In contrast, UV-based methods are proceeded in the unfolded space, which is agnostic to the Euclidean distance.
As shown in Fig.~\ref{fig:ablation_zju_walking_on_triangles}, this results in artifacts. In contrast, our method can eliminate the artifacts by providing more accurate regularization from more accurate triangles.

Moreover, our carefully optimized implementation (Algorithm~\ref{alg:walk_on_mesh}) runs at a neglectable cost of \textbf{1.7} ms over \textbf{200x} faster than the UV-based strategy implemented by SplattingAvatar~\cite{splattingavatar}.

\begin{table}[htbp]
	\centering
	\resizebox{0.9\linewidth}{!}
	{
		\begin{tabular}{c|c|ccc}
			\toprule
			& Time$\downarrow$ & PSNR$\uparrow$ & SSIM$\uparrow$ & LPIPS$\downarrow$\\
			\midrule
			w/o walk  &	-  & 31.07 & 0.9726 & 0.0282 \\
			uv walk	&  208ms	& 31.01 & 0.9724 & 0.0292 \\
			\textbf{Ours} & \textbf{1.7ms}	& \textbf{31.11} & \textbf{0.9728} & \textbf{0.0279} \\
			\bottomrule
		\end{tabular}
	}
	\caption{\textbf{Ablation study on the Walking-on-Mesh strategy.} Mean results over all subjects from ZJU-MoCap is reported.
	}\label{tab:ablation_zju_walking_on_triangles}
\end{table}

\subsubsection{Design of the Adhered-on-Mesh representation}
We conduct an ablation study on the design of the Adhered-on-mesh representation applied in the Adhered Stage by comparing it with the \textbf{fixed-on-mesh} representation from~\cite{gomavatar}. It rigidly binds the Gaussian to fixed location on the mesh, and only optimizes the mesh vertices. We evaluate them on Peoplesnapshot dataset. 

\begin{figure}[htbp]
	\includegraphics[width=1.0\linewidth]{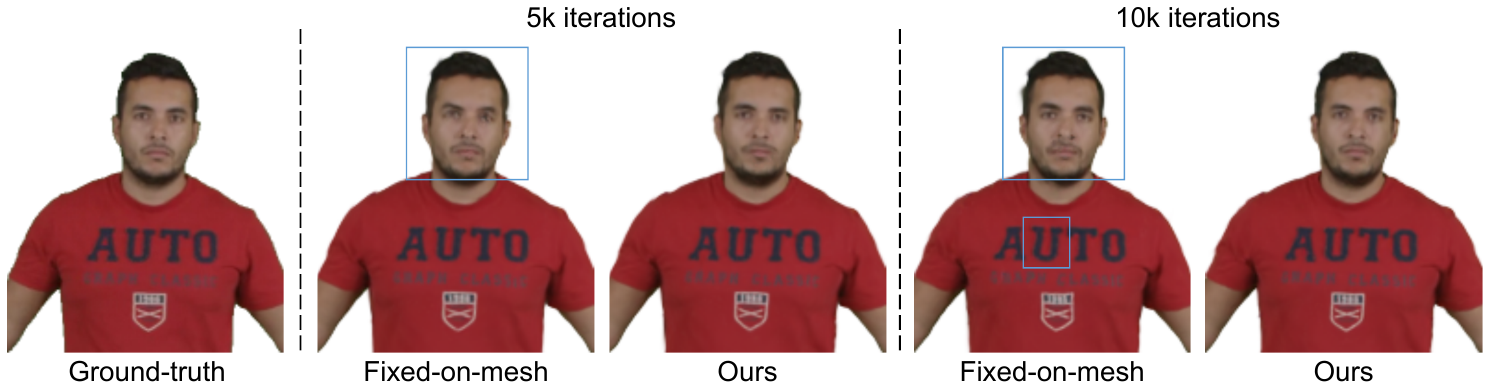}
	\caption{\textbf{Qualitative results of the ablation study on the design of the Adhered-on-mesh representation.} Our flexible representation achieves higher quality in less training time.}\label{fig:ablation_adhered_on_mesh}
\end{figure}

Our approach outperforms the counterparts on all metrics in Tab.~\ref{tab:ablation_ps_adhered_on_mesh}. 
Additionally, as shown in Fig.~\ref{fig:ablation_adhered_on_mesh}, \textbf{ours} can already synthesize photorealistic results after 5k training iterations. In contrast, the \textbf{fixed-on-mesh} representation still struggles with distorted eyes and characters on the chest even after 10k iterations. This demonstrates the flexibility and faster convergence of our adhered-on-mesh representation, yielding more realistic results.
\begin{table}[htbp]
	\centering
	\resizebox{0.75\linewidth}{!}
	{
		\begin{tabular}{c|ccc}
			\toprule
			& PSNR & SSIM & LPIPS\\
			\midrule
			Fixed-on-Mesh  & 32.02 & 0.9787 & 0.0122 \\
			\textbf{Ours} & \textbf{32.19} & \textbf{0.9789} & \textbf{0.0119} \\
			\bottomrule
		\end{tabular}
	}
	\caption{\textbf{Ablation study on the design of the Adhered-on-mesh representation.} Performance is evaluated on the PeopleSnapshot dataset.}\label{tab:ablation_ps_adhered_on_mesh}
\end{table}

\subsection{Applications in geometric reconstruction}
High-quality geometry reconstruction from 3DGS is still under-explored. Existing works typically focus on the static scenes~\cite{sugar}, leaving dynamic human reconstruction from monocular video an unsolved problem. 
Here, we provide a viable solution, and showcase that existing Gaussian-based human avatar methods~\cite{qian20233dgs,gauhuman}, though achieving comparable rendering quality, cannot reconstruct high-quality geometry. On the contrary, leveraging the mesh as a geometry regularizer, our surface-aligned representation significantly improves the geometric quality.

\begin{figure}[htbp]
	\includegraphics[width=1.0\linewidth]{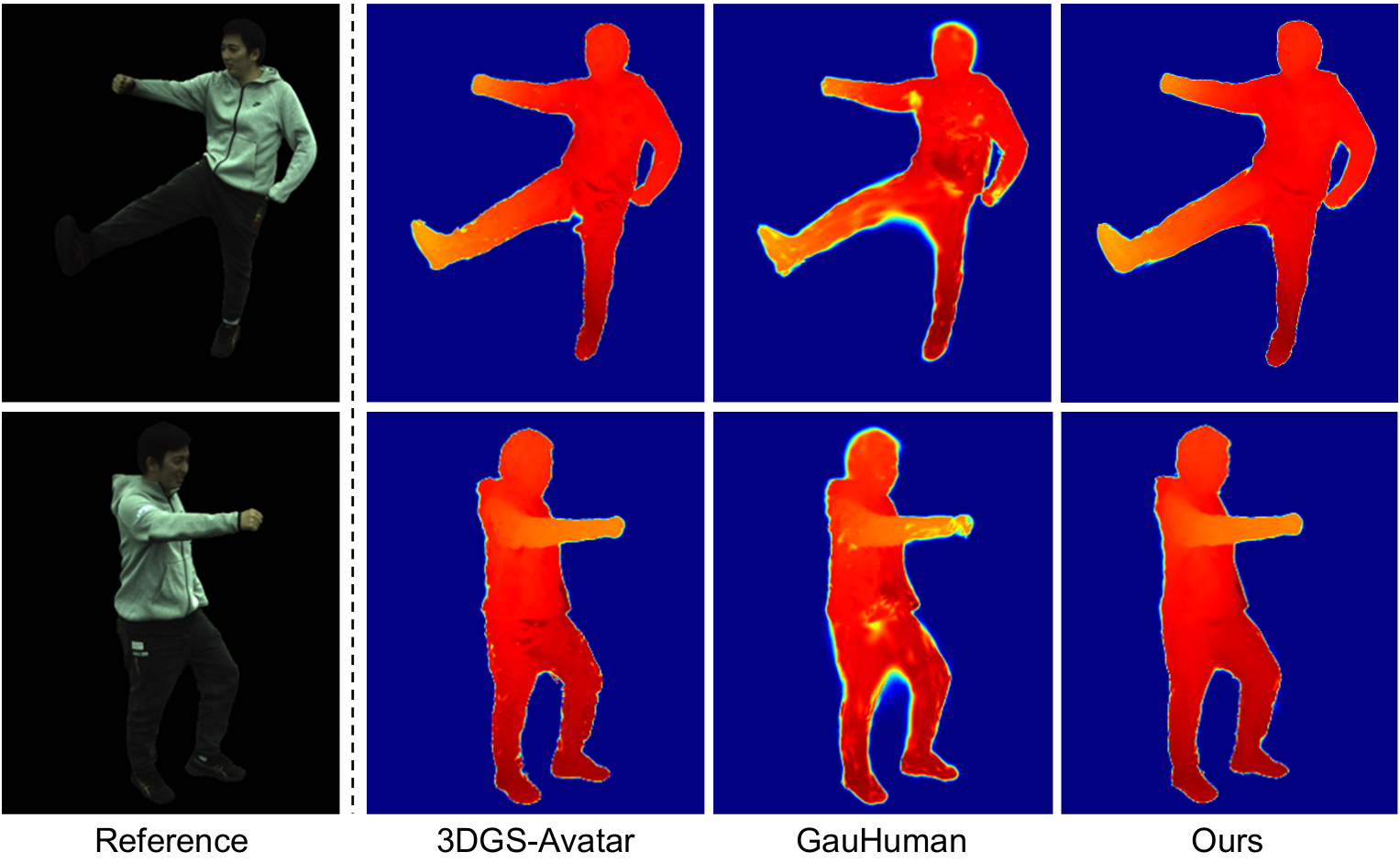}
	\caption{\textbf{
Qualitative comparison of depth rendered by naive Gaussian-based methods~\cite{gauhuman, qian20233dgs}.} While other results suffer from discontinuity and holes, our results are smooth and complete.
}\label{fig:depth}
\end{figure}

\paragraph{Visualization of rendered depth.}
As shown in Fig.~\ref{fig:depth},
naive Gaussian representations~\cite{qian20233dgs, gauhuman} cannot learn faithful geometry, especially in the textureless area, \emph{e.g.}, the pants. This validates our assumption that the monocular input cannot guide 3DGS to form well-defined geometry, leading to severe overfitting. Our method, by contrast, produces more plausible depth images.

\paragraph{Surface reconstruction results.}
We conduct surface reconstruction on subjects from ZJUMocap and Peoplesnapshot datasets by rendering multi-view depth images and fuse them via TSDF fusion~\cite{tsdf}. We sample depth cameras uniformly from a sphere.

\begin{figure}[htbp]
	\includegraphics[width=1.0\linewidth]{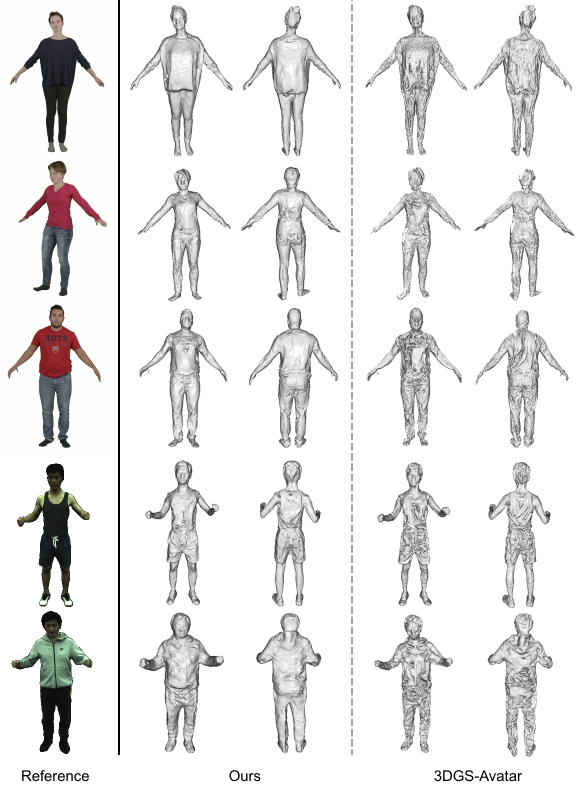}
	\caption{\textbf{Surface reconstruction results.} Our proposed SAGA significantly improves the geometric quality comparing to the SOTA naive Gaussian based method 3DGS-Avatar~\cite{qian20233dgs}.}\label{fig:reconstruction}
\end{figure}

As shown in Fig.~\ref{fig:reconstruction}, naive Gaussian based method 3DGS-Avatar~\cite{qian20233dgs} reconstructs extremely noisy surface, while our method achieves smoother and more accurate results. This validates our surface-aligned representation's ability to generate high-quality geometry and improve generalization under novel views and poses. Furthermore, it demonstrates the potential of our proposed SAGA in Gaussian-based dynamic surface reconstruction.

\section{Conclusion}

This paper presents a two-stage Surface-aligned Gaussian representation for monocular human reconstruction and rendering. In the first stage, the on-mesh Gaussians are allowed to flow, with their centers and normals aligned with the bound triangles. In the second stage, we unleash the fitting ability of Gaussians by detaching them from the mesh, while maintaining the geometry quality by introducing the Gaussian-Mesh Alignment regularization. Additionally, we propose a Walking-on-Mesh algorithm to accurately track the bound triangle as the Gaussians move during optimization. Our method efficiently fits the scene in $\sim$12 minutes and renders in real-time at over 60 FPS.
Extensive experiments on challenging datasets demonstrate that the proposed surface-aligned Gaussian representation effectively regularizes the Gaussians to generate superior geometry without compromising the fitting capabilities of the original 3DGS. This leads to significantly better performance in novel view and novel pose synthesis compared to the state-of-the-art Gaussian-based methods, while marking the first successful attempt at deformable Gaussian-based mesh extraction from monocular videos.

\bibliographystyle{IEEEtran}
\bibliography{main}

\section{Biography Section}
\vspace{-35pt}
\begin{IEEEbiography}[{\includegraphics[width=1in,height=1.25in,clip,keepaspectratio]{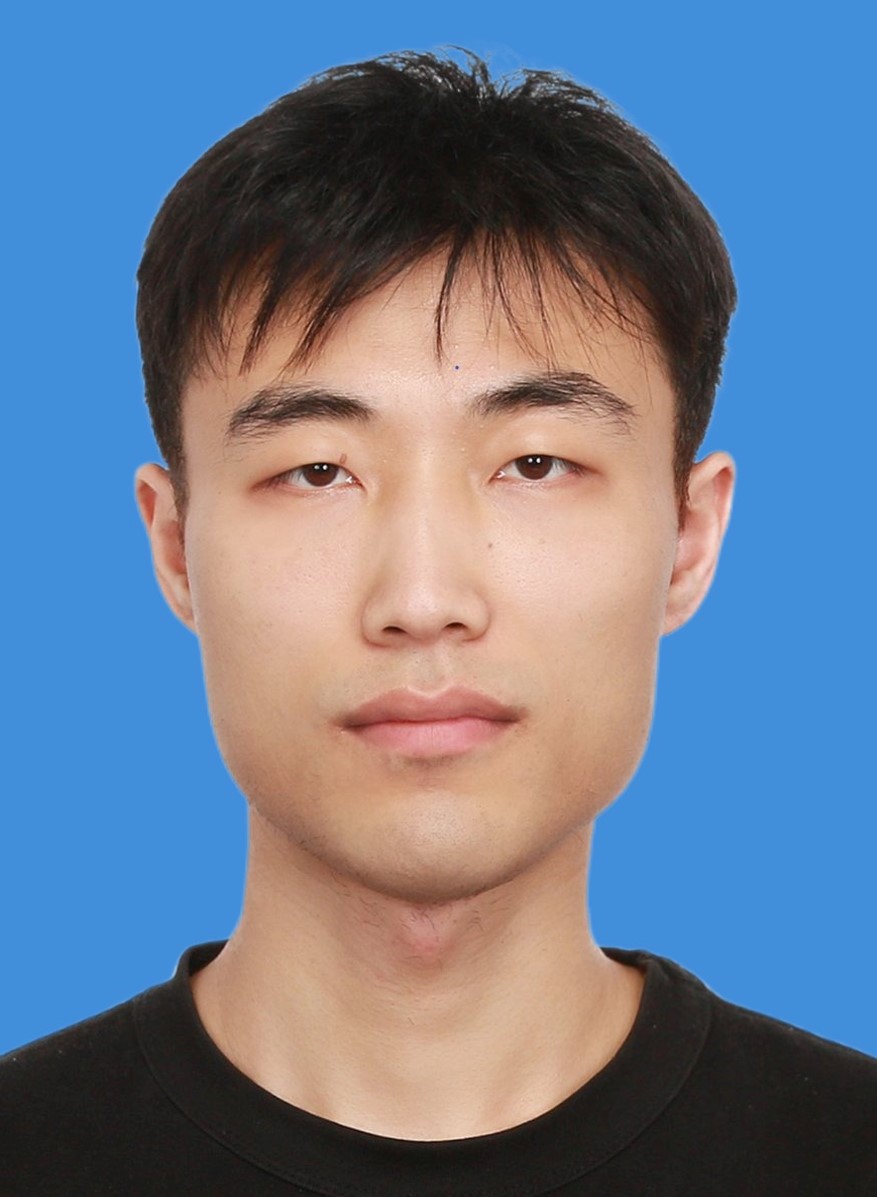}}]{Ronghan Chen}
	Ronghan Chen is currently a Ph.D candidate in State Key Laboratory of Robotics, Shenyang Institute of Automation, Chinese Academy of Sciences. His current research interests include 3D reconstruction and neural rendering.
\end{IEEEbiography}
\vspace{-35pt}
\begin{IEEEbiography}[{\includegraphics[width=1in,height=1.25in,clip]{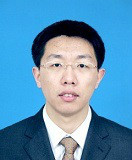}}]{Yang Cong}
	(S’09-M’11-SM’15) is a full professor of the College of Automation Science and Engineering, South China University of Technology. He received the he B.Sc. de. degree from Northeast University in 2004, and the Ph.D. degree from State Key Laboratory of Robotics, Chinese Academy of Sciences in 2009. He was a Research Fellow of National University of Singapore (NUS) and Nanyang Technological University (NTU) from 2009 to 2011, respectively; and a visiting scholar of University of Rochester. He has served on the editorial board of the Journal of Multimedia. His current research interests include image processing, compute vision, machine learning, multimedia, medical imaging, data mining and robot navigation. He has authored over 70 technical papers. He is also a senior member of IEEE.
\end{IEEEbiography}
\vspace{-35pt}
\begin{IEEEbiography}[{\includegraphics[width=1in,height=1.25in,clip,keepaspectratio]{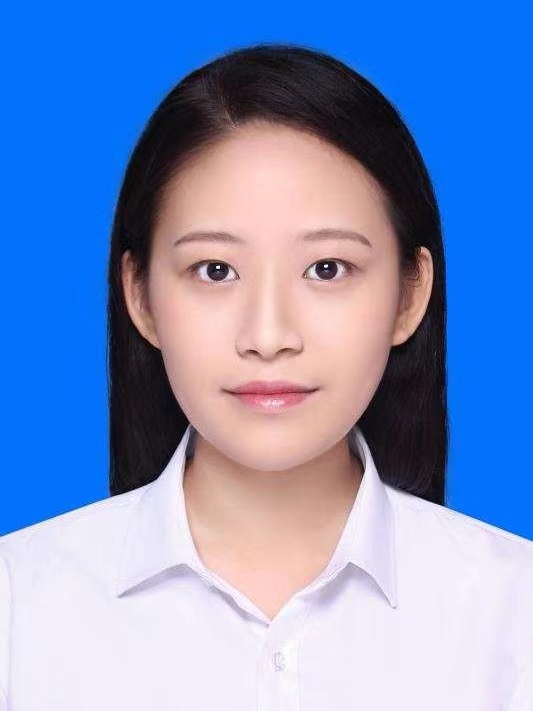}}]{Jiayue Liu}
	Jiayue Liu is a Ph.D candidate in the School of Automation Science and Engineering, South China University of Technology. Her current research interests include 3D reconstruction.
\end{IEEEbiography}

\newpage

\setcounter{section}{0}

\end{document}

%% file: preamble.tex
\usepackage{xcolor}
\usepackage{skak}

\renewcommand{\paragraph}[1]{\vspace{0.5em}\noindent\textbf{#1}}

%% file: pkgs.tex
\usepackage{graphicx}
\usepackage{tikz}
\usepackage{pifont}% http://ctan.org/pkg/pifont
\usepackage{float}
\usepackage{enumitem}
\usepackage{balance}
\usepackage{dsfont}
\usepackage{lipsum}  
\usepackage[thinc]{esdiff}

\usepackage{times}
\usepackage{epsfig}
\usepackage{graphicx}
\usepackage{float}
\usepackage{amsmath}
\usepackage{amssymb}
\usepackage{soul}
\usepackage{multicol}
\usepackage{blindtext}
\usepackage{here}
\usepackage{caption}
\usepackage{makecell}
\usepackage[normalem]{ulem}
\usepackage{multirow}
\usepackage{booktabs}
\usepackage{xcolor, colortbl}
\usepackage{balance}
\usepackage{numprint}
\usepackage{sidecap}